\documentclass[journal,twoside,web]{ieeecolor}
\usepackage{generic}
\usepackage{cite}
\usepackage{amsmath,amssymb,amsfonts}
\usepackage{algorithmic}
\usepackage{graphicx}
\usepackage{textcomp}

\usepackage{pifont}
\usepackage{makecell}
\usepackage{arydshln}
\usepackage{verbatim}
\usepackage{stfloats}
\usepackage{multirow}

\usepackage{threeparttable}
\usepackage{booktabs}

\usepackage[caption=false,font=normalsize,labelfont=sf,textfont=sf]{subfig}
\usepackage[colorlinks,linkcolor=subsectioncolor, anchorcolor=subsectioncolor, urlcolor=subsectioncolor, citecolor=subsectioncolor, allcolors=subsectioncolor]{hyperref}
\newcommand{\figref}[1]{\textcolor{subsectioncolor}{ Fig.~\ref{#1}}}
\newcommand{\tabref}[1]{\textcolor{subsectioncolor}{ Table~\ref{#1}}}

\makeatletter
\newcommand\figcaption{\def\@captype{figure}\caption}
\newcommand\tabcaption{\def\@captype{table}\caption}
\makeatother
\usepackage{array} 

\newsavebox\CBox
\def\textBF#1{\sbox\CBox{#1}\resizebox{\wd\CBox}{\ht\CBox}{\textbf{#1}}}

\def\BibTeX{{\rm B\kern-.05em{\sc i\kern-.025em b}\kern-.08em
		T\kern-.1667em\lower.7ex\hbox{E}\kern-.125emX}}
\begin{document}
\title{3D Vascular Segmentation Supervised by 2D Annotation of Maximum Intensity Projection}
\author{Zhanqiang Guo, Zimeng Tan, Jianjiang Feng, \IEEEmembership{Member, IEEE}, and Jie Zhou, \IEEEmembership{Senior Member, IEEE}
	\thanks{Manuscript received June 11, 2023. This study got ethical approval of  Wuhan Union Hospital of China and Xuanwu Hospital of Capital Medical University (2020009) for using the clinically collected dataset. \emph{(Corresponding author: Jianjiang Feng.)}}
	\thanks{Zhanqiang Guo, Zimeng Tan, Jianjiang Feng, and Jie Zhou are with the Department of Automation, Tsinghua University, Beijing 100084,
		China (e-mail: guozq21@mails.tsinghua.edu.cn; tzm19@mails.tsinghua.edu.cn; jfeng@tsinghua.edu.cn;
		jzhou@tsinghua.edu.cn).}}

\maketitle

\begin{abstract}
	Vascular structure segmentation plays a crucial role in medical analysis and clinical applications. The practical adoption of fully supervised segmentation models is impeded by the intricacy and time-consuming nature of annotating vessels in the 3D space. This has spurred the exploration of weakly-supervised approaches that reduce reliance on expensive segmentation annotations. Despite this, existing weakly supervised methods employed in organ segmentation, which encompass points, bounding boxes, or graffiti, have exhibited suboptimal performance when handling sparse vascular structure. To alleviate this issue, we employ maximum intensity projection (MIP) to decrease the dimensionality of 3D volume to 2D image for efficient annotation, and the 2D labels are utilized to provide guidance and oversight for training 3D vessel segmentation model. Initially, we generate pseudo-labels for 3D blood vessels using the annotations of 2D projections. Subsequently, taking into account the acquisition method of the 2D labels, we introduce a weakly-supervised network that fuses 2D-3D deep features via MIP to further improve segmentation performance. Furthermore, we integrate confidence learning and uncertainty estimation to refine the generated pseudo-labels, followed by fine-tuning the segmentation network. Our method is validated on five datasets (including cerebral vessel, aorta and coronary artery), demonstrating highly competitive performance in segmenting vessels and the potential to significantly reduce the time and effort required for vessel annotation. Our code is available at: \url{https://github.com/gzq17/Weakly-Supervised-by-MIP}.
\end{abstract}


\section{Introduction}
\label{sec:introduction}

Tree-like vascular structures are ubiquitously present within the human body, often characterized by intricate complexities observed at a microscale. Prominent instances of such intricate networks encompass cerebral vessels, aorta, and coronary arteries. Computed Tomography Angiography (CTA) and Magnetic Resonance Angiography (MRA) have emerged as invaluable imaging modalities, facilitating the acquisition of extensive vascular image datasets that have advanced vascular structures research. CTA imaging techniques usually require the injection of contrast agent to highlight blood flow, and it is a contrast-based, minimally invasive, and cost-efficient imaging modality \cite{fu2020rapid}. And MRA techniques rely on blood flow or inflow angiography, augmenting flowing blood's radiance in comparison to stationary tissue through the employment of a short echo time and flow compensation \cite{zhang2020cerebrovascular}. 

The automatic and accurate segmentation of vessels from CTA and MRA is an essential prerequisite in clinical diagnosis and intervention for vascular diseases. Convolutional Neural Networks (CNNs)-based algorithms have demonstrated impressive performance across diverse computer vision tasks, including the segmentation of vascular structures \cite{tetteh2020deepvesselnet,yao2023tag,wang2020deep,jiang2024ori,tan2022retinal}. However, the performance of CNNs is contingent upon large annotated datasets, which are tedious and expensive to obtain, especially for vascular images. Consequently, it is meaningful to develop weakly-supervised methods that leverage weak annotations instead of voxel-wise annotations.

\begin{figure}[t]
	\captionsetup[subfigure]{font={normalfont,scriptsize}}
	\subfloat[\normalfont{Original image}]{\includegraphics[width=.28\linewidth]{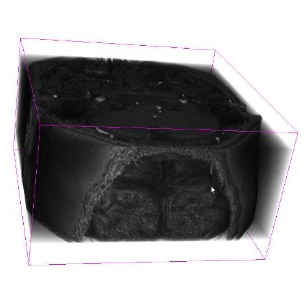}
		\label{pic1_a}%
	}
	\hfil
	\subfloat[\normalfont{Cross-section}]{\includegraphics[width=.28\linewidth]{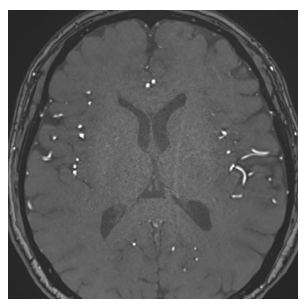}\label{pic1_b}%
	}
	\hfil
	\subfloat[\normalfont{MIP}]{\includegraphics[width=.28\linewidth]{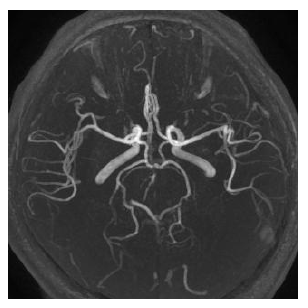}\label{pic1_c}%
	}
	
	\caption{\protect\subref{pic1_a} is an original 3D image. The characteristics of vessels on 2D sections are ambiguous and scattered, as shown in \protect\subref{pic1_b}. \protect\subref{pic1_c} is the MIP image of \protect\subref{pic1_a}. Compared with \protect\subref{pic1_a}, annotating vessels in MIP image is obviously much easier.}
	\label{pic1_intro}
\end{figure}

Various weakly supervised annotations have been used for different types of segmentation tasks, including image-level category labels \cite{li2022online}, bounding boxes \cite{du2023weakly,dorent2021inter}, scribbles \cite{chen2022scribble2d5,zhang2022cyclemix}, and key points \cite{qu2020weakly,guo2021learning}. While these weak annotations demonstrate favorable performance in natural images and large organ segmentation, their utility for sparse blood vessel segmentation remains limited. Image-level annotation is not suitable for segmentation tasks where object classes in images are usually fixed, such as blood vessels and background in our task. The vascular structure typically occupies a small portion of the overall image in the number of voxels, yet exhibits extensive spatial extension, rendering bounding box annotations insufficient in providing substantial informative cues. Scribbles annotation is primarily feasible for 2D images, but given the small size and large number of blood vessels in 2D slices, it is difficult and time-consuming to label, as illustrated in \figref{pic1_intro}\subref{pic1_b}. Key points annotation, such as hundreds of bifurcation points and endpoints of blood vessels, is also laborious to locate and label. Consequently, these weak labels have been scarcely employed in vessel segmentation investigations. Another weakly supervised method for 3D segmentation is to fully annotate a subset of slices within the training volume \cite{wickramasinghe2022weakly,cciccek20163d}. While this approach does alleviate the segmentation burden, the process of annotating these specific slices remains time-intensive, particularly when dealing with vessel slices (\figref{pic1_intro}\subref{pic1_b}). Furthermore, the annotation of 2D slices lacks the essential information pertaining to vascular connectivity, which is crucial for segmentation of 3D vessels. 

Reducing the dimensionality of 3D space to 2D image for annotation and supervision is another intuitive approach in weakly supervised learning. And this has been widely used in 3D point cloud human pose estimation tasks, where the pose information from 2D image is used to aid in the supervision of 3D point cloud pose estimation \cite{kocabas2019self,wu2022c3p,zhang2020weakly}. MRA and CTA techniques harness the principle of blood flow or inflow angiography to impart a brighter appearance to blood vessels relative to surrounding tissue during imaging. Leveraging this property, the maximum intensity projection (MIP) technique projects the maximum voxel value along a specified direction onto the resulting plane \cite{napel1992ct}. By compressing 3D data into 2D projected image, MIP achieves data dimensionality reduction and it is a widely-used scientific method for visualization of vessel structures, vascular analysis, diagnosis, and surgical planning \cite{kiefer2006fast}. In clinical practice, radiologists routinely conduct a swift examination of MIP images to rapidly identify the location, shape, and blood flow characteristics of vessels \cite{salvi2022vision}. MIP images offer abundant information about blood vessels for 3D images, with annotating vessels in MIP images proving significantly easier than directly annotating in 3D space, as depicted in \figref{pic1_intro}\subref{pic1_c}. Regrettably, limited efforts have been made towards utilizing MIP images directly in the study of 3D vessel segmentation. 

In this paper, to alleviate the reliance on 3D vascular annotation and tackle the inapplicability of current weakly supervised labels in 3D vascular segmentation, we introduce a novel approach for weakly supervised learning incorporating the concept of data dimensionality reduction supervision. Leveraging the features unveiled by MIP images, we propose to guide the segmentation of blood vessels in 3D space via 2D MIP annotations. Specifically, we undertake annotation on the MIP image obtained from 3D volume. To utilize effectively of this weak annotation, we back-project the 2D label to generate a sparse 3D foreground image. Subsequently, a region growing algorithm is applied to obtain a more complete labeling of the image. Furthermore, we propose a weakly-supervised segmentation network based on 2D-3D feature fusion to enhance the accuracy of vessel segmentation, taking into account the acquisition method of the 2D labels. Moreover, we integrate the confidence learning (CL) \cite{northcutt2021confident} and uncertainty estimation (UE) via Monte Carlo dropout to further improve the reliability of the generated pseudo-labels, followed by a fine-tuning procedure on the segmentation network to refine its performance. Our contributions can be summarized as follows:

\begin{itemize}
	\item Considering the sparsity of 3D blood vessels, we propose a weakly supervised segmentation framework based on MIP annotations. To the best of our knowledge, this is the first work to utilize MIP image annotations as weakly supervised labels for 3D vessel segmentation.
	\item A segmentation network that fuses 2D-3D features is developed to make full use of designed weak label. And we integrate confident learning and uncertainty estimation to further improve the network's performance.
	\item We validate the effectiveness and generalization of our method on five datasets. Additionally, through carefully designed experiments, we demonstrate that our method achieves superior performance compared to fully supervised segmentation methods while requiring less annotation time when utilizing larger quantities of weakly annotated data.
\end{itemize}

\section{Related Work}
\label{sec:related}

\subsection{Weakly-Supervised Segmentation}
\label{subsec:weak}

Benefits from the development of CNNs, remarkable progress has been accomplished in the field of weakly supervised segmentation. Li et al. \cite{li2022online} implemented an Online Easy Example Mining method for weakly-supervised segmentation of glands using patch-level category labels. Dorent et al. \cite{dorent2021inter} combined the features of extreme points and bounding boxes to supervise the segmentation of vestibular schwannoma and achieved good result. Scribble is obtainable for most segmentation tasks and Zhang et al. \cite{zhang2022cyclemix} adopted the mixup strategy with a dedicated design of random occlusion to perform increments and decrements of scribbles. Meanwhile, as a label for weakly supervised segmentation, key points are often used for object segmentation with regular shape. Guo et al. \cite{guo2021learning} proposed a weakly supervised learning method for nuclei segmentation that required annotation of the nuclear centroid. Nevertheless, due to the sparsity characteristic displayed by 3D vessels, these weak labels (image-level labels, bounding boxes, graffiti and points) are unsuitable for our specific task, as discussed in \textcolor{subsectioncolor}{Sec.} \ref{sec:introduction}.

In practice, most existing weakly supervised approaches of vascular segmentation rely on traditional vessel enhancement techniques to obtain initial segmentation results, which are then iteratively refined manually \cite{vepa2022weakly,zhao2015automated}. Fu et al. \cite{fu2023robust} introduced to supervise the segmentation of LSCI images by choosing best binary labels acquired through various combinations of thresholds. However, the effectiveness and annotation workload of these methods depend on the quality of initial labels. Moreover, since it is time-consuming to correct the labels in 3D images, these methods are mostly used in 2D vessel segmentation. Aiming to 3D hepatic vessel segmentation, Xu et al. \cite{xu2022anti} used both high-quality labeled data and noisy labeled data to train their proposed Mean-Teacher-Assisted network. Nevertheless, different from weakly supervised training, this method still required high-quality annotations, and the outcomes were reliant on the quality of noisy labels. In this work, we utilize the annotation of 2D MIP image to supervise the vessel segmentation in 3D space, which greatly reduces the annotation time. 

\subsection{Noisy Pseudo-Label Refinement}
\label{subsec:refinement}

The primary step in weakly- and semi-supervised learning is generating pseudo-labels of training data, leveraging weakly labeled or pre-existing fully labeled data. To improve the robustness of trained models, recent studies have focused on refining the noisy pseudo-labels. Particularly, uncertainty estimation has also emerged as a common approach for optimizing noisy labels \cite{wang2021uncertainty,yang2021uncertainty}. For instance, Cao et al. \cite{cao2020uncertainty} estimated uncertainty to discern potential noise in the generated pseudo-labels, consequently mitigating the detrimental impact on network performance. Additionally, Northcutt et al. \cite{northcutt2021confident} proposed a confident learning method capable of identifying potentially incorrect samples in noisy labels through uncertainty estimation, subsequently removing them during training. This approach is gradually being adopted for optimizing generated noisy pseudo-labels \cite{xu2022anti,guo2021learning}. However, these techniques exclusively address already labeled data. And, the generated pseudo-labels via MIP labels used in our work only cover a part of the voxels. Consequently, we employ the confidence learning method to refine the already-labeled voxels during the pseudo-label refinement, while simultaneously integrating the uncertainty estimation to assign labels to unlabeled voxels. 

\subsection{Dimensionality Reduction Supervision and MIP}
\label{subsec:mip}

The utilization of annotation information derived from low-dimensional data to enhance analysis in high-dimensional spaces has found widespread application across various domains, especially in the field of 3D point cloud pose estimation \cite{kocabas2019self,wu2022c3p,zhang2020weakly}. Zhang et al. \cite{zhang2020weakly} employed adversarial learning to leverage weakly supervised data comprising solely annotations of 2D human joints, enabling the recovery of human pose. Similarly, Wu et al. \cite{wu2022c3p} introduced a refined point set network structure to transfer annotation information obtained from 2D human pose estimation within existing large-scale RGB datasets to the 3D task.

For 3D CTA and MRA images, MIP is an intuitive method for dimensionality reduction, projecting 3D voxels onto a projection plane based on their maximum intensity. MIP images prove valuable in facilitating rapid observation of vascular structures and blood flow characteristics by medical professionals. Salvi et al. \cite{salvi2022vision} trained a vision transformer using MIP images for the diagnosis of peripheral arterial disease. And recent studies have emphasized the combination of MIP image features to enhance algorithmic performance when analyzing 3D images. Chen et al. \cite{chen20233d} leveraged prior knowledge demonstrating the similarity in tree structures between 2D and 3D blood vessels, employing an adversarial learning method to utilize existing 2D blood vessel annotations to supervise the fidelity of the MIP image of the 3D segmentation result. And Kozi{\'n}ski et al. \cite{kozinski2020tracing,kozinski2018learning} combined 3D-to-2D cross entropy loss and multiple annotated projection images to reduce the annotation effort required for 3D linear structure segmentation. However, the use of projection information is considerably constrained in these studies. For instance, in the study by Dima et al. \cite{dima20233d}, the reliance on preprocessing steps and the limitations imposed by the type of vascular tissue and imaging method influenced the utilization of projection images. Furthermore, some researchers have utilized projection information to enhance the learning of image features \cite{oner2022enforcing,zheng2019automatic}. For instance, Zheng et al. \cite{zheng2019automatic} utilized MIP images with varying plate thicknesses as input to augment the spatial information of CT images and aid in discriminating between nodules and blood vessels. Wang et al. \cite{wang2020jointvesselnet} integrated MIP image embedding into 3D MRA to extract vessel structures. However, these studies primarily employed MIP to extract feature and required complete 3D vascular annotations. In our work, we employ MIP technique to achieve dimensionality reduction of 3D volume to 2D image, which serves the purpose of facilitating annotation and supervision. The proposed method brings about a remarkable reduction in the required annotation time, while simultaneously guaranteeing the quality of vessel segmentation. 

\section{Method}
\label{sec:method}

\begin{figure*}[t]
	\centerline{\includegraphics[width=17cm]{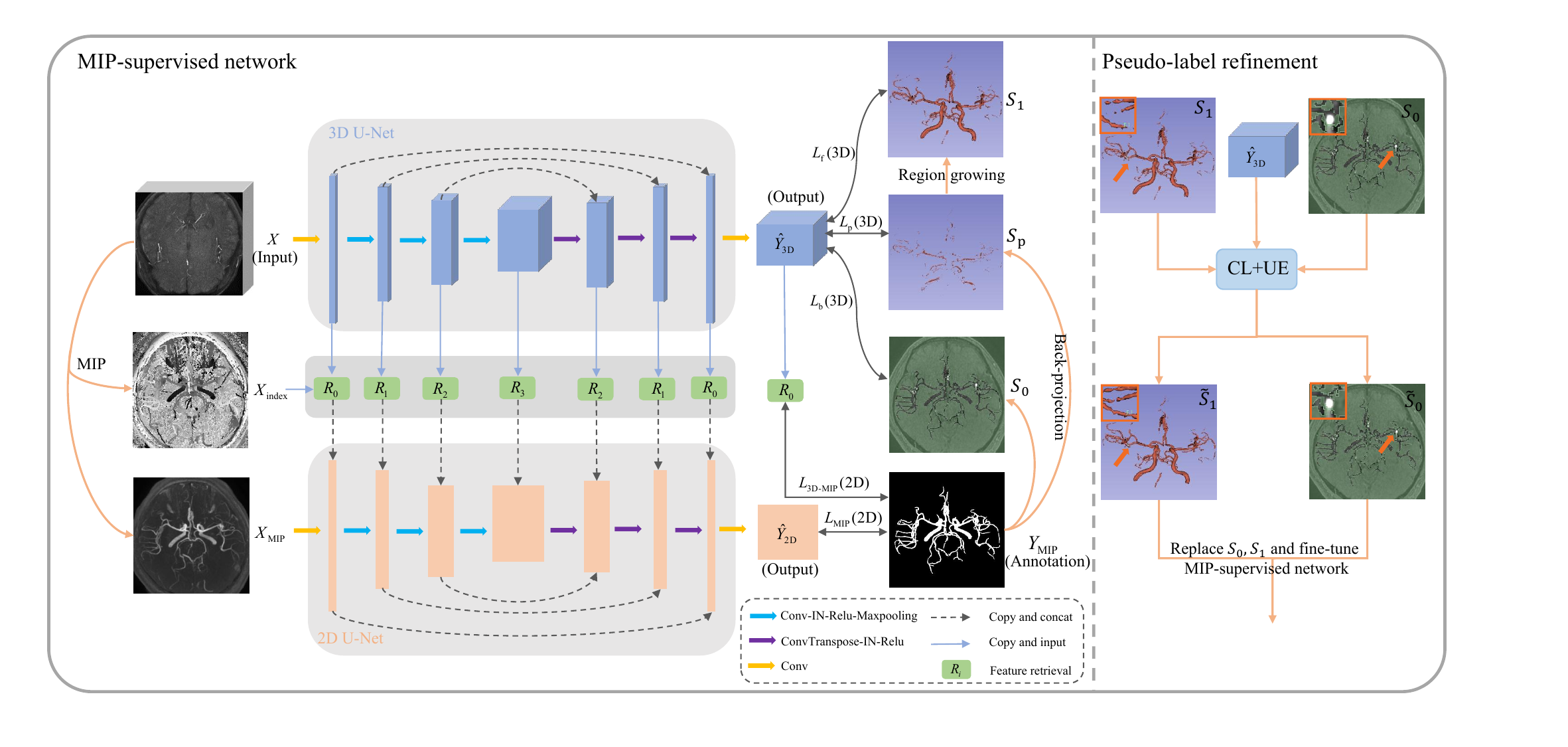}}
	\caption{The proposed weakly-supervised vessel segmentation framework. Initially, we employ MIP to reduce dimensionality of 3D volume for annotating. Subsequently, we introduce a novel 2D-3D feature fusion network, which is trained with pseudo labels generated from 2D annotations. To enhance the efficacy of the network, we integrate confidence learning and uncertainty estimation methods to refine the pseudo labels, followed by fine-tuning of the network.}
	\label{pic2}
\end{figure*}

The overall pipeline of the proposed weakly-supervised vascular segmentation framework is illustrated in \figref{pic2}. Due to the limited number of pixels in the 2D MIP labels, we first generate 3D pseudo-labels based on them. Subsequently, the proposed 2D-3D feature fusion network is trained with 2D weak annotations and the newly generated 3D pseudo-labels. To enhance the credibility of the pseudo-labels, confidence learning in combination with uncertainty estimation is employed to optimize the labels, followed by fine-tuning of the network. Each step of the proposed methodology will be described in detail in this section.

\subsection{MIP and 3D Pseudo-labels Generation}
\label{subsec:psedo-label}

Let $X \in R^{H\times W\times D}$ denote a 3D image and $\Omega = \{(x,y,z)\}^{H\times W\times D}$ denote the set of all points in 3D space. In the following description, for convenience, we project the image in the transverse plane. We perform MIP of $X$ to obtain the 2D projection image $X_{\mathrm{MIP}}$ and index map $X_{\mathrm{index}}$, mathematically expressed as $X_{\mathrm{MIP}}(x,y)=\max\limits_{z=1,2..,D}X(x,y,z)$ and $X_{\mathrm{index}}(x,y)=\mathop{\arg\max}\limits_{z=1,2..,D}X(x,y,z)$. $Y_{\mathrm{MIP}}$ is the vessel annotation of projection image, a weak label of 3D image $X$. However, $Y_{\mathrm{MIP}}$ is extremely sparse for 3D volume, so it is necessary to generate credible 3D pseudo-label to train the segmentation model. Using the index map, we back-project the labeled 2D image to get a series of discrete points $S_{\mathrm{p}}$ in 3D space, which are foreground voxels. The back-project operation is expressed as:
\begin{equation}
	S_{\mathrm{p}} = \{(x,y,z);\;Y_{\mathrm{MIP}}(x,y)=1, X_{\mathrm{index}}(x,y)=z\}.
\end{equation}

To increase the amount of foreground voxels for supervision, we treat $S_{\mathrm{p}}$ as seed points and employ a region growing algorithm to obtain the set $S_1$. The algorithm starts from the seed points $S_{\mathrm{p}}$ and gradually adds adjacent voxels to the foreground until the predefined stopping criterion is reached. The criterion we choose is that the difference between the voxel value of the candidate point and the average voxel value of seed points is lower than a preset threshold $\alpha$.

A straightforward approach of generating the background voxels set is to utilize all columns in the 3D image that correspond to the background points in the 2D projection label, denoting as $T_{\mathrm{b1}}=\{(x,y,z); Y_{\mathrm{MIP}}(x,y)=0, z=1,2,...,D\}$. However, two facts are ignored: (i) Most voxels on the columns correspond to foreground pixels in the 2D label are also background, but are not included in $T_{\mathrm{b1}}$. These voxels are denoted as $T_{\mathrm{b2}}=\{(x,y,z);X(x,y,z)<\beta_{\mathrm{th}}$ $\mathrm{and}$ $ Y_{\mathrm{MIP}}(x,y)=1\}$; (ii) During projection, certain blood vessels may be obstructed by brighter noise, resulting in being displayed as background in 2D projection image. $T_{\mathrm{b3}}=\{(x,y,z);\gamma_{\mathrm{th}} <X(x,y,z)<\eta_{\mathrm{th}}$ $\mathrm{and}$ $ Y_{\mathrm{MIP}}(x,y)=0\}$ represents these voxels. So the corrected background voxels set $S_{\mathrm{0}}$ is computed as $S_{\mathrm{0}} = T_{\mathrm{b1}} \cup T_{\mathrm{b2}} / T_{\mathrm{b3}}$.

The parameters $\beta_{\mathrm{th}}$, $\gamma_{\mathrm{th}}$ and $\eta_{\mathrm{th}}$ are related to the average gray value $v_{\mathrm{ave}}$ of the known foreground, computed as $v_{\mathrm{ave}}=\frac{1}{|S_{\mathrm{p}}|}\sum_{p\in S_{\mathrm{p}}}X(p)$. In our experiment, $\beta_{\mathrm{th}}=0.2v_{\mathrm{ave}}$, $\gamma_{\mathrm{th}}=1.2v_{\mathrm{ave}}$, $\eta_{\mathrm{th}} = 1.6v_{\mathrm{ave}}$.

\subsection{Weakly-Supervised Network}
\label{subsec:network}

To effectively leverage the information provided by the weak annotation, a 2D-3D deep feature fusion network is designed based on U-Net \cite{ronneberger2015u} and 3D U-Net \cite{cciccek20163d}, denoted as $g_{\mathrm{2D}}$ and $g_{\mathrm{3D}}$ respectively, as shown in \figref{pic2}. In fact, most fully supervised methods can also serve as backbone for our scheme. The 3D image $X$ and the corresponding 2D projection image $X_{\mathrm{MIP}}$ are fed into two networks to obtain the prediction probability maps, $\hat{Y}_{\mathrm{3D}}=g_{\mathrm{3D}}(X)$, $\hat{Y}_{\mathrm{2D}}=g_{\mathrm{2D}}(X_{\mathrm{MIP}})$.

The use of MIP image effectively captures the spatial information, geometric attribute, and interconnectivity of 3D blood vessels. Moreover, the availability of ground truth supervision for MIP image enhances the reliability of $g_{\mathrm{2D}}$ prediction, so it is important to flow the information from $g_{\mathrm{2D}}$ to $g_{\mathrm{3D}}$ during training. Within our proposed network framework, we establish a linkage between the feature map of two networks based on the inherent relationship exhibited by their respective inputs. This facilitates utilization of the segmentation information generated by the 2D network within the 3D network. Specifically, the features extracted from two networks are connected using the index map $X_{\mathrm{index}}$ obtained during MIP. For the extracted 3D feature of the $i$-th layer $f_i^{\mathrm{3D}}\in R^{C\times \frac{H}{2^i}\times \frac{W}{2^i} \times \frac{D}{2^i}}$, the corresponding 2D feature $f_i^{\mathrm{3D}\mapsto \mathrm{2D}}\in R^{C\times \frac{H}{2^i}\times \frac{W}{2^i}}$ is calculated by the feature retrieval module $R_i$ as follows:
\begin{equation}
	\begin{aligned}
	f_i^{\mathrm{3D}\mapsto \mathrm{2D}}(c,x,y)&=R_i(f_i^{\mathrm{3D}})\\ &=f_i^{\mathrm{3D}}(c,x,y,X_{\mathrm{index}}^i(x,y)),\\ & i\in\{0,1,2,3\},
	\end{aligned}
\label{eqR}
\end{equation}
where $X_{\mathrm{index}}^i\in R^{\frac{H}{2^i}\times \frac{W}{2^i}}$ is obtained by interpolating the index map $\frac{X_{\mathrm{index}}}{2^i}$. And the feature $f_i^{\mathrm{3D}\mapsto \mathrm{2D}}$ is concatenated with the corresponding feature layer of the 2D U-Net.

\subsection{Confident Learning and Uncertainty Estimation}
\label{subsec:clandun}

Two challenges arise in the pseudo-labels ($S_0, S_1$) generated in \textcolor{subsectioncolor}{Sec.} \ref{subsec:psedo-label}: (i) the pseudo-label generation process, which relies on seed points and grayscale information, unavoidably introduces noise into the labels; (ii) the pseudo-labels offer coverage only for a subset of voxels, whereas many voxels remain unlabeled ($|S_0\cup S_1|<|\Omega|$). To tackle these issues, we propose the incorporation of confidence learning (CL) and uncertainty estimation (UE) to further refine the pseudo-labels.

\subsubsection{Noisy Labeled Voxels Refinement with CL}
To identify and address the presence of noisy labels within the pre-labeled voxels ($\Omega_{\mathrm{L}}=S_0\cup S_1$), the true (latent) foreground and background sets are estimated using the network output ($\hat{Y}_{\mathrm{3D}}$):
\begin{equation}
	S_i^*=\{p; p\in \Omega_{\mathrm{L}}, i=\mathop{\arg\max}\limits_{j}^{}\hat{y}_{\mathrm{3D}}^p(j), \hat{y}^p_{\mathrm{3D}}(i) > t_i\},
\end{equation}
where $t_i$ is average self-confidence of the labeled set $S_i$, that is $t_i=\frac{1}{|S_i|}\sum_{q\in S_i}\hat{y}_{\mathrm{3D}}^q(i)$. And $\hat{y}_{\mathrm{3D}}^p(i)$ is the  predicted probability ($\hat{Y}_{\mathrm{3D}}$) belonging to the $i$-th category at point $p$. And then we calculate the normalized count matrix $\tilde{C}_{S,S^*}$ as:
\begin{equation}
	\tilde{C}_{S, S_{}^{*}}[i][j]=\frac{|S_i\cap S_j^*|}{\mathop{\sum}\limits_{j\in \{0,1\}}|S_i\cap S_j^*|}\cdot |S_i|,
\end{equation}
where the reason for normalization is $|S_0^*\cup S_1^*|\leq |\Omega_\mathrm{L}|$ affected by the threshold $t_i$. Subsequently, we estimate the joint distribution based on $\tilde{C}_{S,S^*}$:
\begin{equation}
	\hat{Q}_{S, S_{}^{*}}[i][j]=\frac{\tilde{C}_{S, S_{}^{*}}[i][j]}{\mathop{\sum}\limits_{i\in \{0,1\}}\mathop{\sum}\limits_{j\in \{0,1\}}\tilde{C}_{S, S_{}^{*}}[i][j]}.
\end{equation}

The mislabeled voxels set (the set to remove) is selected by $\hat{Q}_{S, S_{}^{*}}[i][j]$ with the Prune by Noise Rate (PBNR) strategy \cite{northcutt2021confident}, expressed as:
\begin{equation}
	\begin{aligned}
	S_i^{\mathrm{(re)}}=&\{p;\, y^p_{\mathrm{3D}}(1-i)-y^p_{\mathrm{3D}}(i)>i_{\mathrm{th}},p\in S_{i}\cap S_{1-i}^*\}\\
	&\cup (S_0 \cap S_1),
	\end{aligned}
\end{equation}
where $i_{\mathrm{th}}$ is the minimum value of the top $|\Omega_{\mathrm{L}}|\cdot 	\hat{Q}_{S, S_{}^{*}}[i][1-i]$ in the set $\{y^q_{\mathrm{3D}}(1-i)-y^q_{\mathrm{3D}}(i);q\in S_{i}\cap S_{1-i}^*\}$. And, the voxels that exist simultaneously in $S_0$ and $S_1$ are also removed. We select some of wrongly labeled background voxels to add to the foreground set based on prior knowledge, and vice versa:
\begin{equation}
	S_i^{\mathrm{(add1)}}=\{p;\,\,p\in S_{1-i}^{\mathrm{(re)}},\,\, \mathrm{and}\,\mathrm{prior}\, \mathrm{condition} \},
	\label{CL_computing}
\end{equation}
where the \textit{prior condition} is $X(p) < \varepsilon_1 v_{\mathrm{ave}}$ when $i=0$, while $D(p, S_1) < d_{\mathrm{th1}}$ when $i=1$. And $D(p, S_1)$ is the minimum distance from point $p$ to set $S_1$.

\subsubsection{Unlabeled Voxels Refinement with UE} 
To address the issue of unlabeled voxels ($\Omega_{\mathrm{U}}=\Omega/\Omega_{\mathrm{L}}$), we adopt uncertainty estimation method to assign labels to reliable voxels. This process begins with the measurement of uncertainty for each voxel, utilizing the Monte Carlo dropout method. For each training data $X$, we execute multiple forward passes ($K$ times, $K=6$ in our experiments) using $g_{3D}$ with dropout to obtain prediction probabilities $\{\hat{Y}_k\}_{k=1}^{K}$:
\begin{equation}
	\hat{Y}_k = g_{3D}(X+\mathcal{N}_k(\mu, \sigma^2)),
\end{equation}
where $\mathcal{N}_k(\mu, \sigma^2)$ is a stochastic Gaussian distribution with mean $\mu$ and variance $\sigma^2$, with the dimensions matching those of $X$ ($\mu=0$, $\sigma=0.1$ in our experiments). Meanwhile, we compute prediction probability $\hat{Y}=g_{\mathrm{3D}}(X)$ without dropout and the probability result with dropout $\hat{Y}_{\mathrm{dp}}=\frac{1}{K}\mathop{\sum}_{k}\hat{Y}_k$. Subsequently, the uncertainty of each voxel is computed as:
\begin{equation}
	u_p=-\mathop{\sum}\limits_{i\in \{0,1\}}(\frac{1}{K}\mathop{\sum}\limits_{k}\hat{y}_k^p(i))\cdot \mathrm{log}_2(\frac{1}{K}\mathop{\sum}\limits_{k}\hat{y}_k^p(i)).
\end{equation}
Finally, based on the uncertainty, we determine the additional set of foreground and background points:
\begin{equation}
	\begin{aligned}
		u_{\mathrm{ave}}^i=\frac{1}{\mathop{\sum}\limits_{p\in\Omega_{\mathrm{U}}}[y^p=y^p_{\mathrm{dp}}=i]}\cdot \mathop{\sum}\limits_{p\in\Omega_{\mathrm{U}}}[y^p=y^p_{\mathrm{dp}}=i]\cdot u_p,
	\end{aligned}
\end{equation}
\begin{equation}
	\begin{aligned}
	S_{i}^{\mathrm{(add2)}}=\{p;&\,p\in\Omega_{\mathrm{U}},y^p=y^p_{\mathrm{dp}}=i,\, u_p<u_{\mathrm{ave}}^i,\\& \mathrm{and}\,\mathrm{prior}\, \mathrm{condition}\},
	\end{aligned}
	\label{Un_computing}
\end{equation}
where $y^p=\mathop{\arg\max}\limits_{j}^{}\hat{y}^p(j)$,  $y^p_{\mathrm{dp}}=\mathop{\arg\max}\limits_{j}^{}\hat{y}^p_{\mathrm{dp}}(j)$, $[\cdot]$ is indicator function. And the \textit{prior condition} is $X(p) < \varepsilon_2 v_{\mathrm{ave}}$ when $i=0$, while $D(p, S_1) < d_{\mathrm{th2}}$ when $i=1$.

The foreground and background sets after refinement are calculated as:
\begin{equation}
	\begin{aligned}
		\tilde{S}_i=(S_i\cup S_i^{\mathrm{(add1)}}/S_i^{\mathrm{(re)}})\cup S_{i}^{\mathrm{(add2)}}.
	\end{aligned}
\end{equation}

\subsection{Loss Function}
\label{subsec:loss}

As shown in \figref{pic2}, the loss function consists of two components, $L_{\mathrm{2D}}$ and $L_{\mathrm{3D}}$. The output of 2D U-Net is under supervision via MIP annotation, whereas the output of the 3D network is under the guidance of 3D pseudo-labels. As the 3D pseudo-labels do not cover all the voxels, we employ a weighted cross-entropy loss expressed as:
\begin{equation}
	\begin{aligned}
	L_{\mathrm{3D}}&= L_{\mathrm{f}}(\mathrm{3D})+L_{\mathrm{p}}(\mathrm{3D})+L_{\mathrm{b}}(\mathrm{3D})\\
	&=-\frac{1}{|S_{\mathrm{f}}|} \sum_{p\in S_{\mathrm{f}}}^{} \mathrm{log}(\hat{y}_{\mathrm{3D}}^{p}(1))-\frac{1}{|S_{\mathrm{p}}|}\sum_{p\in S_{\mathrm{p}}}^{} \mathrm{log}(\hat{y}_{\mathrm{3D}}^{p}(1))\\
	&\quad -\frac{1}{|S_{\mathrm{b}}|}\sum_{p\in S_{\mathrm{b}}}^{} \mathrm{log}(\hat{y}_{\mathrm{3D}}^p(0)),
	\end{aligned}
\end{equation}
where $S_{\mathrm{f}}=S_{\mathrm{1}}$ and $S_{\mathrm{b}}=S_{\mathrm{0}}$ in the initial training phase, while $S_{\mathrm{f}}=\tilde{S}_{\mathrm{1}}$ and $S_{\mathrm{b}}=\tilde{S}_{\mathrm{0}}$ during fine-tuning the network. As $Y_{\mathrm{MIP}}$ serves as the ground truth for $X_{\mathrm{MIP}}$, this component is supervised by the Dice loss, a commonly used loss function for segmentation task:
\begin{equation}
	\begin{aligned}
	L_{\mathrm{2D}}&=L_{\mathrm{3D\text{-}MIP}}(\mathrm{2D})+L_{\mathrm{MIP}}(\mathrm{2D})\\
	&=\mathrm{Dice}(R_0(\hat{Y}_{\mathrm{3D}}), Y_{\mathrm{MIP}}) +  \mathrm{Dice}(\hat{Y}_{\mathrm{2D}}, Y_{\mathrm{MIP}}),
	\end{aligned}
\end{equation}
where $R_0(\cdot)$ is defined in the \textcolor{subsectioncolor}{Eq.} \ref{eqR}. The final loss $L_{\mathrm{all}}$ is expressed as:
\begin{equation}\fontsize{9.5}{5}
	L_{\mathrm{all}}=L_{\mathrm{3D}}+\lambda L_{\mathrm{2D}}.
\end{equation}

\section{Experiments}
\label{sec:experiment}

\subsection{Datasets and Preprocess}
\label{subsec:dataset}

We evaluate our method on five datasets, including three cerebrovascular datasets, a coronary CTA dataset, and an aortic CTA dataset.

\subsubsection{TubeTK} 
The publicly available dataset TubeTK\footnote{\underline{https://public.kitware.com/Wiki/TubeTK/Data}} comprises 42 3D time-of-flight MRA volumes with labeled vessels (centerline + radius). To facilitate further analysis, we convert the annotation into voxel data using the MetaIO\footnote{\underline{https://itk.org/Wiki/MetaIO}}.

\subsubsection{Cerebral MRA}
This dataset consists of 96 MRA cerebrovascular volumes acquired from various imaging systems. The images are retrospectively collected from Xuanwu Hospital of Capital Medical University, China. Each sample has 3D vessel annotation. During annotating, we employ Frangi filtering \cite{frangi1998multiscale} to generate the initial segmentation result for the vessels. Then, fine corrections are made by two radiologists to obtain the final vascular label.

\subsubsection{Cerebral CTA}
It comprises 47 3D CTA cerebrovascular volumes with blood vessel annotations. The source and labeling process of this dataset are consistent with those of the Cerebral MRA dataset. However, for CTA data, the skull region is highlighted, affecting the segmentation of blood vessels and causing obstruction during MIP. To mitigate this issue, the CTA data has been performed skull-stripping to remove the bright skull regions \cite{muschelli2015validated}.

\subsubsection{Coronary CTA}
\begin{figure}[h]
	\captionsetup[subfigure]{font={tiny}}
	\subfloat[]{\includegraphics[width=.2\linewidth]{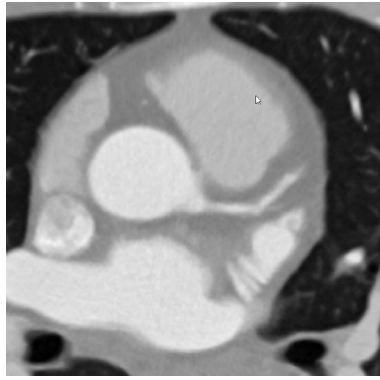}
		\label{pic3_a}%
	}
	\subfloat[]{\includegraphics[width=.2\linewidth]{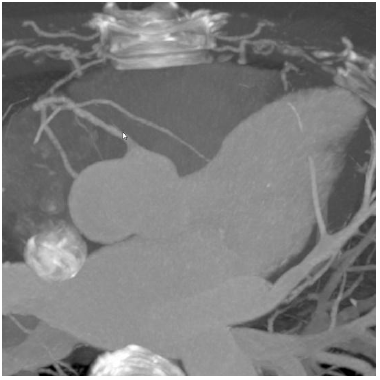}\label{pic3_b}%
	}
	\hfil
	\subfloat[]{\includegraphics[width=.2\linewidth]{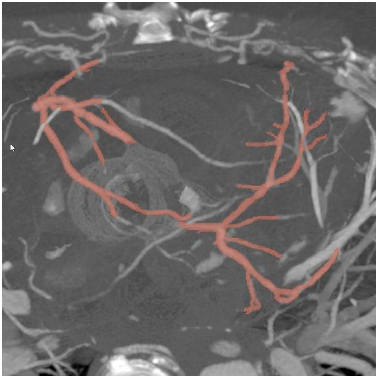}\label{pic3_c}%
	}
	\hfil
	\subfloat[]{\includegraphics[width=.165\linewidth]{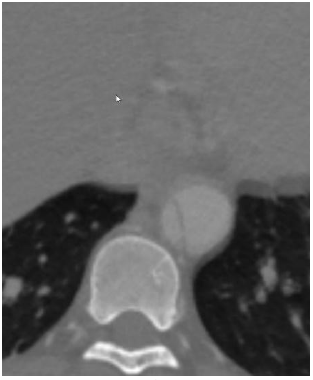}\label{pic3_d}%
	}
	\hfil
	\subfloat[]{\includegraphics[width=.075\linewidth]{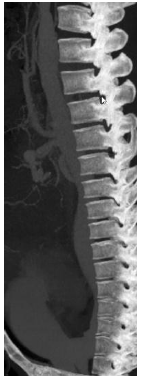}\label{pic3_e}%
	}
	\hfil
	\subfloat[]{\includegraphics[width=.075\linewidth]{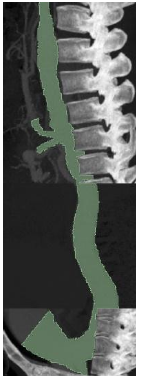}\label{pic3_f}%
	}
	\caption{Preprocessing of Coronary CTA and Aorta CTA datasets: \protect\subref{pic3_a} and \protect\subref{pic3_d} depict a slice of the coronary and aorta volume, respectively. \protect\subref{pic3_b} and \protect\subref{pic3_e} show the MIP images obtained through direct projection, revealing that a significant portion of vessels is obscured by other anatomical structures. The MIP images of processed volumes are displayed in \protect\subref{pic3_c} and \protect\subref{pic3_f}, with a clear display of the majority of the blood vessel.}
	\label{pic3_preprocess}
\end{figure}

The coronary dataset contains 52 3D CTA volumes, which are retrospectively collected from Wuhan Union Hospital of China. And the annotation of coronary arteries is completed by a radiologist. The grayscale value of the ascending aorta, left atrium, and other some parts is found to be higher than that of the coronary artery, resulting in occlusion during MIP. To address this, we apply a region growing method to remove the ascending aorta, and subsequently employ a combination of the threshold method and region growing method for each slice to remove the remaining high-intensity areas, as shown in \figref{pic3_preprocess}. Notably, this step is solely performed during the MIP process of training samples, and the original image served as the input during training. So no processing steps is needed during inference.

\subsubsection{Aorta CTA}
This dataset comprises 50 aorta volumes from Wuhan Union Hospital of China. Annotation of the aorta is done by a radiologist. In contrast to other datasets mentioned above, the unique structure of aorta limits the utility of MIP image projected in transverse plane. Therefore, we perform MIP in the sagittal plane on aorta dataset. Similar to Coronary CTA dataset, the aorta is also affected by other brighter parts during projection, primarily concentrated in the middle layers. And we employ a traditional algorithm based on shape prior \cite{biesdorf2009automatic} to process the middle part of layers and remove possible occluded regions (in experiments, we process the middle 230-380 layers, as shown in \figref{pic3_preprocess}). This step is, again, exclusive to the MIP process.

\begin{table}[h]
	\caption{The resolution (Spacing) and size (Cropping) of adjusted data. The number of train/validation/test sets (Number) and whether some parts were removed (Removing).}
	\scriptsize
	\centering
	\label{tab1}\renewcommand\arraystretch{1.15}
	\begin{tabular}{ c  c  c  c  c}
		\Xhline{2\arrayrulewidth}
		\textbf{Dataset} & \textbf{Spacing(mm$^3$)} & \textbf{Cropping} & \textbf{Number} & \textbf{Removing} \\
		\hline
		TubeTK & 0.5$\times$0.5$\times$0.8 & 384$\times$ 384$\times$128  & 30/4/8 & \ding{56}\\
		Cerebral MRA & 0.5$\times$0.5$\times$0.75 & 320$\times$ 320$\times$128  & 30/4/62 & \ding{56}\\
		Cerebral CTA & 0.5$\times$0.5$\times$0.75 & 320$\times$ 320$\times$128  & 30/5/12 & \ding{51}\\
		Coronary CTA & 0.4$\times$0.4$\times$0.4 & 320$\times$ 320$\times$256 & 30/4/18 & \ding{51} \\
		Aorta CTA & 1.0$\times$1.0$\times$1.0 & 128$\times$ 160$\times$480 & 30/4/16 & \ding{51} \\
		\Xhline{2\arrayrulewidth}
	\end{tabular}
\end{table}

The resolution of the data in each dataset is standardized and subsequently the volumes are cropped to ensure consistent size, leaving the middle vessel area. In cases where the image size is insufficient, zero padding is employed to achieve uniform data size. Additionally, a gray value normalization step is applied, mapping the intensity range to 0-1. The images of each dataset are randomly partitioned into training, validation, and test sets. \tabref{tab1} provides an overview of the adjusted data, encompassing information on resolution, data size, the specific allocation of images, as well as any pre-processing steps executed to eliminate potential occlusions within the data that might impede blood vessel visibility during MIP.

\begin{figure}[b]
	\captionsetup[subfigure]{font={sf,tiny}}
	\subfloat[\normalfont{3D label}]{\includegraphics[width=.24\linewidth]{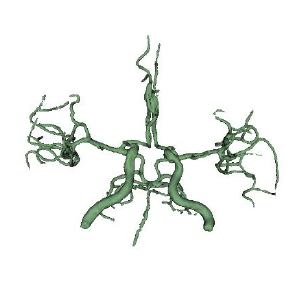}
		\label{pic4_a}%
	}
	\hfil
	\subfloat[\normalfont{Label of 2D slice}]{\includegraphics[width=.24\linewidth]{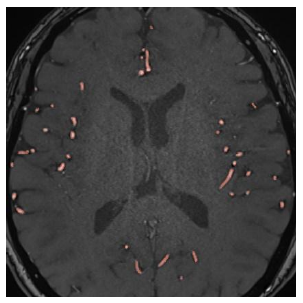}\label{pic4_b}%
	}
	\hfil
	\subfloat[\normalfont{Noisy label}]{\includegraphics[width=.24\linewidth]{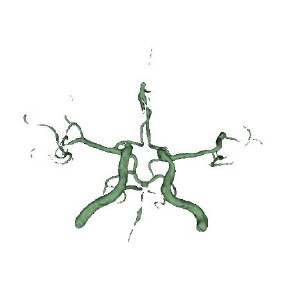}\label{pic4_c}%
	}
	\hfil
	\subfloat[\normalfont{MIP label}]{\includegraphics[width=.24\linewidth]{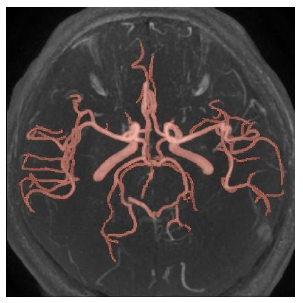}\label{pic4_e}%
	}
	\caption{Annotations of different categories.}
	\label{pic4_annotation}
\end{figure}

\subsection{Metrics and Implementation Details}
\label{subsec:Implementation}

\subsubsection{Metrics}
We utilize the following metrics to evaluate our method: Dice Similarity Coefficient (DSC), ClDice \cite{shit2021cldice}, which is tailored to evaluate tubular structures while accounting for vascular connectivity, and Average Hausdorff Distance (AHD) \cite{beauchemin1998hausdorff}, which incorporates voxel localization considerations \cite{taha2015metrics}. Furthermore, to indicate the statistical significance of improvements of the proposed method, we also present the p-values for DSC using a paired t-test with each comparison method.

\begin{figure*}[b]
	\centerline{\includegraphics[width=17.5cm]{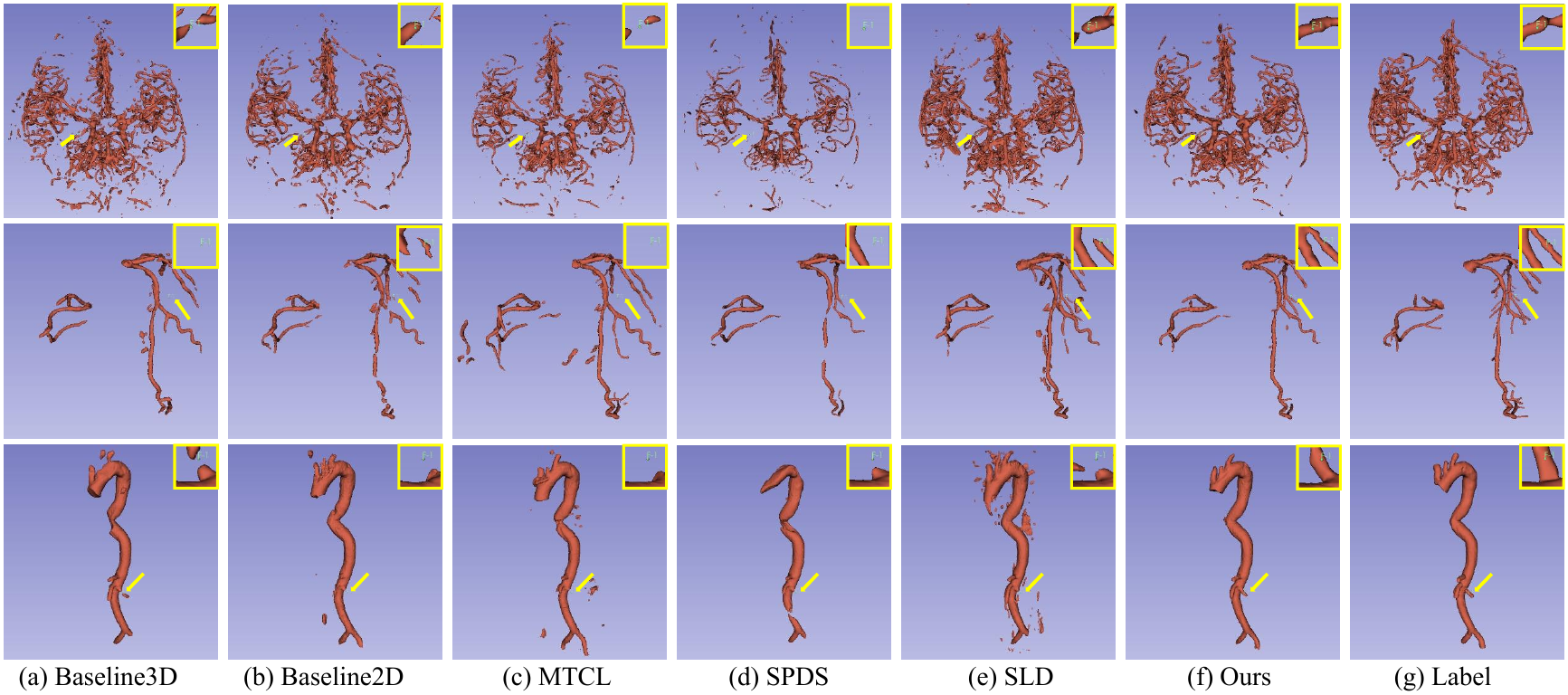}}
	\caption{Segmentation results on three testing images from three datasets (TubeTK, Coronary CTA, Aorta CTA dataset in order). The red boxes highlight close-ups of some vessels.}
	\label{pic5_3d_result}
\end{figure*}

\subsubsection{Implementation Details}
In our proposed weakly supervised network, we employ 3D U-net \cite{cciccek20163d} and 2D U-net \cite{ronneberger2015u} architectures as the backbone. The down-sampling path of two models features convolution layers with filter numbers of [8, 16, 32, 64]. The implementation of the network is conducted using the PyTorch framework. Training process is performed on a NVIDIA GeForce GTX 3090 GPU with 24G memory. During the training process, we utilize the adaptive moment estimation (Adam) optimizer, initialized with a learning rate of 0.001. A decay factor of 0.9 is applied to the learning rate after each iteration. The maximum number of training iterations is set to 1000. We employ the preset parameter $\alpha=0.1$ in pseudo-labels generation by region growing. And the prior parameters $d_{\mathrm{th1}}$, $d_{\mathrm{th2}}$, $\varepsilon_1$ and $\varepsilon_2$ are respectively set to 1.5, 4.0, 0.7 and 0.2 in \textcolor{subsectioncolor}{Eq.} \ref{CL_computing} and \textcolor{subsectioncolor}{Eq.} \ref{Un_computing}. Importantly, it should be noted that when working with the public dataset (TubeTK dataset), the priori information is not set during pseudo-labels refinement. In other words, the parameters $d_{\mathrm{th1}}$, $d_{\mathrm{th2}}$, $\varepsilon_1$ and $\varepsilon_2$ are considered $\infty$. This adjustment is made due to the noisy labels in TubeTK dataset, where the annotated vessels are thin and incorporate some venous structures \cite{hilbert2020brave}. Consequently, during training with pseudo-labels, the TubeTK dataset has a higher tolerance for noise in the labels. Additionally, the balance parameter $\lambda$ in the loss function is set to 1.0. We will shortly make our code publicly available.

\subsection{Compared Methods and Annotation Time}
\label{subsec:methodtime}

\subsubsection{Compared Methods}

\begin{table}[b]
	\scriptsize
	\centering
	\caption{The type of annotated data (Annotation), the number of annotations (Number), the average annotation time (Ave), and the total annotation time of the training samples (All).}
	\label{tab2}\renewcommand\arraystretch{1.15}
	\begin{tabular}{ c  c  c  c c }
		\Xhline{2\arrayrulewidth}
		\textbf{Method} & \textbf{Ave (min)} & \textbf{All (min)} & \textbf{Number}  & \textbf{Annotation}  \\
		\hline
		Full-sup & 73.41 & 2202.30 & $N=30$ & \figref{pic4_annotation}\subref{pic4_a} \\
		\hdashline
		Baseline3D \cite{cciccek20163d} & 73.41 & 220.23  & $m_1=3$ & \figref{pic4_annotation}\subref{pic4_a}\\
		Baseline2D & 10.07 & 302.10  & $s_1=10$ & \figref{pic4_annotation}\subref{pic4_b}\\
		MTCL \cite{xu2022anti} & - & 237.82  & $m_2=3$ & \figref{pic4_annotation}\subref{pic4_a} and \subref{pic4_c}\\
		SLD \cite{chen20233d} & 6.78 & 203.40 & $N=30$ & \figref{pic4_annotation}\subref{pic4_e}\\
		SPDS \cite{dima20233d} & 6.78 & 203.40 & $N=30$  & \figref{pic4_annotation}\subref{pic4_e}\\
		Ours & 6.78 & 203.40 & $N=30$ & \figref{pic4_annotation}\subref{pic4_e}\\
		\Xhline{2\arrayrulewidth}
	\end{tabular}
\end{table}

\begin{table*}[h]
	\centering
	\scriptsize
	\caption{Comparison with other methods on three cerebrovascular datasets, with the best performance highlighted in bold. The $p$ of $DSC(p)$ represents the p$\text{-}$value calculated by the t$\text{-}$test, and $\ast$ indicates the statistical difference between Ours and other methods. ($\ast :p<0.05$, $\ast\ast :p<0.01$, $\ast\ast\ast :p<0.001$)}
	\label{tab3}\renewcommand\arraystretch{1.15}
	\begin{threeparttable}
		\begin{tabular}{ c  c  c  c  c  c  c  c  c  c}
			\Xhline{2\arrayrulewidth}
			\multirow{2}*[-3pt]{\textbf{Method}}
			\rule{0pt}{2pt} & \multicolumn{3}{c}{\multirow{1}{*}[-1pt]{\textbf{\textbf{TubeTK}}}}
			& \multicolumn{3}{c}{\multirow{1}{*}[-1pt]{\textbf{\textbf{Cerebral MRA}}}}
			& \multicolumn{3}{c}{\multirow{1}{*}[-1pt]{\textbf{\textbf{Cerebral CTA}}}} \\ [-1.5pt]
			\cmidrule(lr){2-4}\cmidrule(lr){5-7}\cmidrule(lr){8-10}
			& \multirow{1}{*}[0.8pt]{\textbf{DSC(\%) (p)}} & \multirow{1}{*}[0.8pt]{\textbf{ClDice(\%)}}  & \multirow{1}{*}[0.8pt]{\textbf{AHD(mm)}} 
			& \multirow{1}{*}[0.8pt]{\textbf{DSC(\%)}} & \multirow{1}{*}[0.8pt]{\textbf{ClDice(\%)}}  & \multirow{1}{*}[0.8pt]{\textbf{AHD(mm)}} 
			& \multirow{1}{*}[0.8pt]{\textbf{DSC(\%)}} & \multirow{1}{*}[0.8pt]{\textbf{ClDice(\%)}}  & \multirow{1}{*}[0.8pt]{\textbf{AHD(mm)}}    \\
			\hline
			\multirow{1}{*}{Full-sup} 
			& 64.52 ($\ast\ast\ast$) & 77.60 & 0.917 
			& 85.07 ($\ast\ast\ast$) & 89.00 & 0.303 
			& 85.34 ($\ast\ast\ast$) & 88.92 & 0.288 \\
			\hdashline
			\multirow{1}{*}{Baseline3D (2016) \cite{cciccek20163d}} 
			& 55.39 ($\ast\ast\ast$) & 62.04 & 1.602 
			& 79.81 ($\ast\ast\ast$) & 71.96 & 1.011 
			& 78.18 ($\ast\ast\ast$) & 79.96 & 0.486  \\
			\multirow{1}{*}{Baseline2D}  
			& 59.09 ($\ast\ast\ast$) & 74.11 & 1.029 
			& 82.28 ($\ast\ast\ast$) & 85.40 & 0.436 
			& 81.19 ($\ast\ast\ast$) & 81.80 & 0.406 \\
			\multirow{1}{*}{MTCL (2022)} \cite{xu2022anti}
			& 60.95 \hspace{0.01cm}(0.127)\hspace{0.01cm} & 74.89 & 1.017 
			& 76.01 ($\ast\ast\ast$) & 73.04 & 1.502 
			& 72.34 ($\ast\ast\ast$) & 66.40 & 1.487 \\
			\multirow{1}{*}{SLD (2023)} \cite{chen20233d}
			& 58.40 ($\ast\ast\ast$) & 71.08 & 1.237 
			& 81.09 ($\ast\ast\ast$) & 83.06 & 0.453 
			& 79.77 ($\ast\ast\ast$) & 77.68 & 0.641 \\
			\multirow{1}{*}{SPDS (2023)} \cite{dima20233d} 
			& 55.12 ($\ast\ast\ast$) & 69.14 & 1.794 
			& 79.93 ($\ast\ast\ast$) & 82.31 & 0.426 
			& 78.75 ($\ast\ast\ast$) & 77.35 & 0.476 \\
			\multirow{1}{*}{Ours} 
			& \textbf{61.10} \hspace{0.22cm}(-)\hspace{0.23cm} & \textbf{75.93} & \textbf{0.843} 
			& \textbf{84.35} \hspace{0.22cm}(-)\hspace{0.23cm} & \textbf{87.40} & \textbf{0.336} 
			& \textbf{83.84} \hspace{0.22cm}(-)\hspace{0.23cm} & \textbf{83.10} & \textbf{0.255} \\
			\Xhline{2\arrayrulewidth}
		\end{tabular}
	\end{threeparttable}
\end{table*}

As shown in \tabref{tab1}, our proposed method utilize a training set size of $N=30$ for all five datasets, under supervision via the annotation of 2D MIP image (as shown in \figref{pic4_annotation}\subref{pic4_e}). To illustrate the effectiveness of our proposed method, we implement the following algorithms for comparison.

\begin{itemize}
	\item \textBF{Full-sup:} We report the results of fully supervised model training with $N$ images as the upper bound. The model structure of Full-sup is consistent with our proposed method.
	\item \textBF{Baseline3D \cite{cciccek20163d}:} We use $m_1$ ($m_1 < N$) fully labeled samples (\figref{pic4_annotation}\subref{pic4_a}) to train a fully supervised model, whose structure is consistent with 3D Unet used in our proposed method.
	\item \textBF{Baseline2D:} For each sample, $s_1\%$ layers are randomly chosen and labeled vessel in 2D slice (\figref{pic4_annotation}\subref{pic4_b}). The $N$ volumes with these annotations are then used to train the 3D Unet.
	\item \textBF{MTCL \cite{xu2022anti}:} This method represents a semi-supervised approach for blood vessel segmentation that relies on a small number of fully labeled data and a larger amount of noisy labeled data. Specifically, we apply the Frangi vessel enhancement method \cite{frangi1998multiscale} and get a noisy label (\figref{pic4_annotation}\subref{pic4_c}) by only manually removing obvious noise due to the arduous task of adding blood vessels. And we employ $m_2$ fully labeled data and $N-m_2$ noisy labeled data for MTCL.
	\item \textBF{SLD \cite{chen20233d}:} To ensure fair comparison, we employ the supervision of the MIP annotation as an alternative to the supervision of the adversarial learning component within SLD. And the number of training images is consistent with our approach ($N$).
	\item \textBF{SPDS \cite{dima20233d}:} Similar to SLD \cite{chen20233d}, to ensure methodological consistency, we align the supervision labels employed in SPDS with our approach, and the 3D Unet structure is applied as the backbone in SPDS.

\end{itemize}

\subsubsection{Annotation Time of Each Method}
For a fair comparison, we endeavor to maintain consistency between the annotation time of the training data employed by the compared methods and the annotation time of $N$ MIP images used in our proposed method. We present the average time of labeling data and the total annotation time (taken by a radiologist to manually annotate the images) of training samples for each method on Cerebral MRA, as shown in \tabref{tab2}.

\subsection{Comparative Results}
\label{subsec:comparative}

\begin{table*}[!h]
	\centering
	\scriptsize
	\caption{Comparison with other methods on Coronary CTA and Aorta CTA datasets, with the best performance highlighted in bold. The $p$ of $DSC(p)$ is consistent with \tabref{tab3}.}
	\label{tab4}\renewcommand\arraystretch{1.15}
	\begin{threeparttable}
		\begin{tabular}{ c c c c c c c }
			\Xhline{2\arrayrulewidth}
			\multirow{2}*[-2.8pt]{\fontsize{6pt}{7pt} \textbf{Method}}
			\rule{0pt}{2pt} & \multicolumn{3}{c}{\multirow{1}{*}[-0.6pt]{\fontsize{6pt}{7pt} \textbf{Coronary CTA}}}
			& \multicolumn{3}{c}{\multirow{1}{*}[-0.6pt]{\fontsize{6pt}{7pt} \textbf{Aorta CTA}}} \\ [-1.5pt]
			\cmidrule(lr){2-4}\cmidrule(lr){5-7}
			& \multirow{1}{*}[0.6pt]{\fontsize{6pt}{7pt}\selectfont \textbf{DSC(\%)}} & \multirow{1}{*}[0.6pt]{\fontsize{6pt}{7pt}\selectfont \textbf{ClDice(\%)}}  & \multirow{1}{*}[0.6pt]{\fontsize{6pt}{7pt}\selectfont\textbf{AHD(mm)}} 
			& \multirow{1}{*}[0.6pt]{\fontsize{6pt}{7pt}\selectfont \textbf{DSC(\%)}} & \multirow{1}{*}[0.6pt]{\fontsize{6pt}{7pt}\selectfont \textbf{ClDice(\%)}}  & \multirow{1}{*}[0.6pt]{\fontsize{6pt}{7pt}\selectfont\textbf{AHD(mm)}}   \\
			\hline
			Full-sup 
			& 77.40 \hspace{0.17cm}($\ast\ast$)\hspace{0.13cm} & 75.88 & 0.745 
			& 91.75 \hspace{0.01cm}(0.052)\hspace{0.01cm} & 89.20 & 0.339\\
			\hdashline
			Baseline3D (2016) \cite{cciccek20163d}
			& 67.42 ($\ast\ast\ast$) & 67.23 & 1.706 
			& 87.85 \hspace{0.17cm}($\ast\ast$)\hspace{0.13cm} & 86.21 & 0.698 \\
			 Baseline2D  
			& 73.59 \hspace{0.17cm}($\ast\ast$)\hspace{0.13cm} & \textbf{73.95} & 0.686 
			& 90.47 \hspace{0.25cm}($\ast$)\hspace{0.19cm} & 86.69 & 0.367 \\
			 MTCL (2022) \cite{xu2022anti} 
			& 67.77 \hspace{0.17cm}($\ast\ast$)\hspace{0.13cm} & 64.36 & 2.093
			& 87.05 \hspace{0.17cm}($\ast\ast$)\hspace{0.13cm} & 82.18 & 1.185 \\
			SLD (2023) \cite{chen20233d}
			& 70.45 ($\ast\ast\ast$) & 68.17 & 1.792 
			& 86.81 ($\ast\ast\ast$) & 83.50 & 0.919 \\
			SPDS (2023) \cite{dima20233d}
			& 69.46 ($\ast\ast\ast$) & 68.93 & 1.053 
			& 83.38 ($\ast\ast\ast$) & 79.26 & 1.185 \\
			Ours 
			& \textbf{75.43} \hspace{0.22cm}(-)\hspace{0.23cm} & 73.17& \textbf{0.660} 
			& \textbf{90.84} \hspace{0.22cm}(-)\hspace{0.23cm} & \textbf{91.25} & \textbf{0.348} \\
			\Xhline{2\arrayrulewidth}
		\end{tabular}
	\end{threeparttable}
\end{table*}

\subsubsection{Quantitative results}
We present the quantitative results of the five datasets in \tabref{tab3} and \tabref{tab4}. Under approximately consistent annotation time, our proposed method leverages the annotation of MIP image to provide the model with enhanced information regarding blood vessel direction and connectivity, thereby yielding superior outcomes. The weakly supervised labels proposed in our method offer effective information for blood vessel segmentation while significantly reducing the annotation workload. And the proposed approach effectively harnesses the information, resulting in compelling results that closely approach the performance achieved through full supervision. MTCL \cite{xu2022anti} achieves performance comparable to our proposed method on the TubeTK dataset, potentially due to the resemblance between the noisy labels employed in MTCL and the provided annotations, which contain noise itself \cite{hilbert2020brave}. Notably, the methods SLD \cite{chen20233d} and SPDS \cite{dima20233d} employ the same weak labels as our method to promote fair comparison; however, their utilization of these labels is severely restricted, leading to suboptimal performance. Utilizing identical labels, our proposed framework integrates the projection label acquisition method with the design of the 2D-3D feature fusion network, while also optimizing the resultant pseudo labels, resulting in superior performance. 

\begin{figure*}[t]
	\centerline{\includegraphics[width=17.5cm]{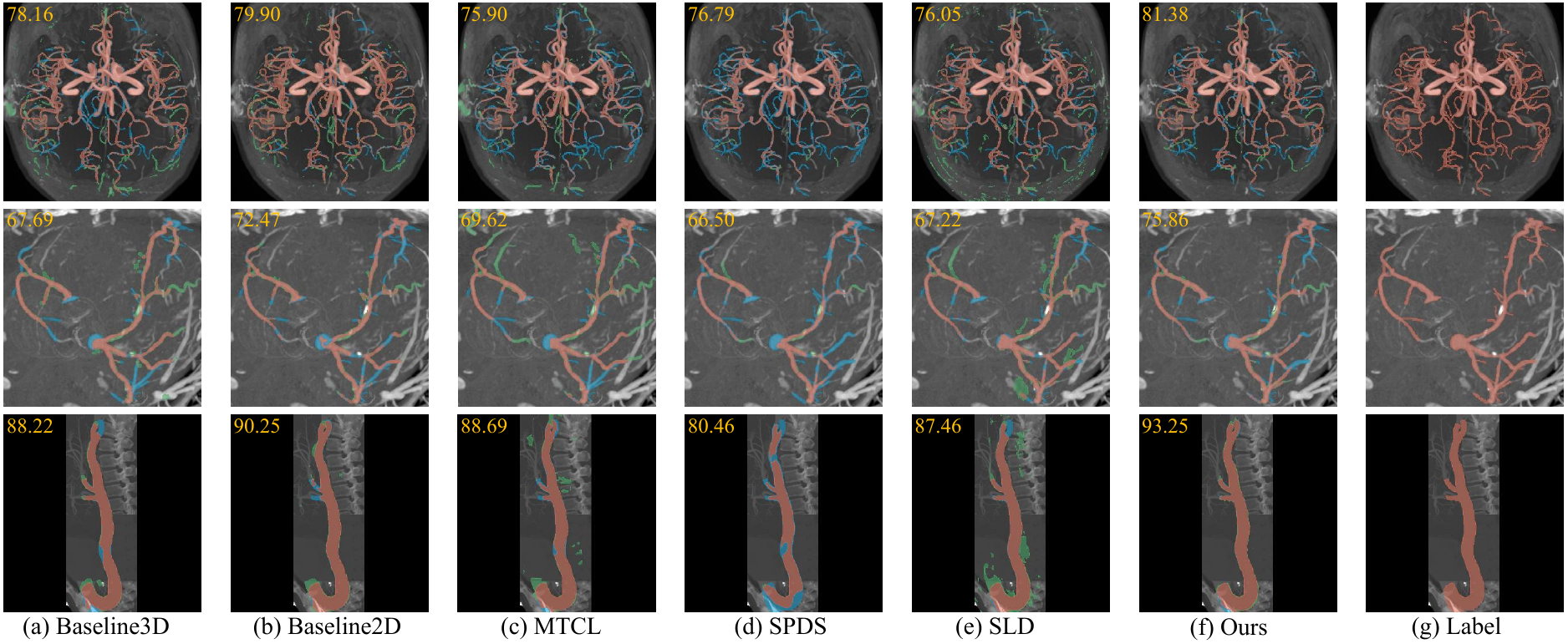}}
	\caption{The MIPs of images and	segmentation results. The first to third rows represent the MIP images corresponding to the segmentation results of one testing sample from the Cerebral MRA, Coronary CTA, and Aorta CTA datasets, respectively. The red pixels, blue pixels and green pixels denote true positives, false negatives and false positives respectively. The Dice score (\%) of corresponding 3D segmentation result is shown in the upper left corner of MIP image.}
	\label{pic6_mip_result}
\end{figure*}

\begin{table*}[t]
	\centering
	\scriptsize
	\caption{The results of abation study. We demonstrate the effectiveness of each component. The $p$ of $DSC(p)$ is consistent with \tabref{tab3}.}
	\label{tab5}\renewcommand\arraystretch{1.15}
	\begin{tabular}{ c  c  c   c  c   c  c   c  c  c  c}
		\Xhline{2\arrayrulewidth}
		
		\multirow{2}*[-3pt]{\textbf{Method}} & \multicolumn{2}{c}{\multirow{1}{*}[-1pt]{\textbf{TubeTK}}} & \multicolumn{2}{c}{\multirow{1}{*}[-1pt]{\textbf{Cerebral MRA}}} & \multicolumn{2}{c}{\multirow{1}{*}[-1pt]{\textbf{Cerebral CTA}}} & \multicolumn{2}{c}{\multirow{1}{*}[-1pt]{\textbf{Coronary CTA}}} & \multicolumn{2}{c}{\multirow{1}{*}[-1pt]{\textbf{Aorta CTA}}} \\ [-1.5pt]
		\cmidrule(lr){2-3}\cmidrule(lr){4-5}\cmidrule(lr){6-7}\cmidrule(lr){8-9}\cmidrule(lr){10-11}
		& \multirow{1}{*}[0.5pt]{\textbf{DSC(\%)}}   & \multirow{1}{*}[0.5pt]{\textbf{AHD(mm)}} & \multirow{1}{*}[0.5pt]{\textbf{DSC(\%)}}   & \multirow{1}{*}[0.5pt]{\textbf{AHD(mm)}} & \multirow{1}{*}[0.5pt]{\textbf{DSC(\%)}}   & \multirow{1}{*}[0.5pt]{\textbf{AHD(mm)}} & \multirow{1}{*}[0.5pt]{\textbf{DSC(\%)}}   & \multirow{1}{*}[0.5pt]{\textbf{AHD(mm)}} & \multirow{1}{*}[0.5pt]{\textbf{DSC(\%)}}   & \multirow{1}{*}[0.5pt]{\textbf{AHD(mm)}}\\
		\hline
		$L_{\mathrm{2D}}$ & 55.39 ($\ast\ast\ast$) & 1.602 
		& 72.73 ($\ast\ast\ast$) & 1.680 
		& 44.35 ($\ast\ast\ast$) & 5.471 
		& 72.24 ($\ast\ast\ast$) & 1.044 
		& 88.71 ($\ast\ast\ast$) & 0.650 \\
		$L_{\mathrm{3D}}$ & 58.01 ($\ast\ast\ast$) & 0.870 
		& 81.66 ($\ast\ast\ast$) & 0.601 
		& 80.52 ($\ast\ast\ast$) & 0.349 
		& 73.63 \hspace{0.17cm}($\ast\ast$)\hspace{0.13cm} & 0.783 
		& 87.34 ($\ast\ast\ast$) & 0.615 \\
		$L_{\mathrm{all}}$ & 60.08 \hspace{0.17cm}($\ast\ast$)\hspace{0.13cm} & 0.971 & 83.74 ($\ast\ast\ast$) & 0.375 
		& 83.53 \hspace{0.25cm}($\ast$)\hspace{0.19cm} & 0.289 
		& 74.17 \hspace{0.25cm}($\ast$)\hspace{0.19cm} & 0.699 
		& 89.90 \hspace{0.25cm}($\ast$)\hspace{0.19cm} & 0.460 \\
		$L_{\mathrm{all}}$+RF & \textbf{61.10} \hspace{0.28cm}(-)\hspace{0.20cm} & \textbf{0.843} & \textbf{84.35} \hspace{0.22cm}(-)\hspace{0.23cm} & \textbf{0.336} & \textbf{83.84} \hspace{0.28cm}(-)\hspace{0.22cm} & \textbf{0.255} & \textbf{75.43} \hspace{0.28cm}(-)\hspace{0.20cm} & \textbf{0.660} & \textbf{90.84} \hspace{0.28cm}(-)\hspace{0.20cm} & \textbf{0.348} \\
		\Xhline{2\arrayrulewidth}
	\end{tabular}
\end{table*}

\subsubsection{Qualitative results}
\figref{pic5_3d_result} exhibits the 3D results of testing data from different datasets. Our proposed method yields better connectivity and more accurate boundary detection of small vessels when compared to other algorithms. These findings are consistent with the results in \tabref{tab3} and \tabref{tab4}. Additionally, the outcome of Baseline2D on Aorta CTA dataset is comparable to that of our proposed approach on metrics of DSC and AHD. This can be attributed to the relatively simple structure of the aorta in comparison to coronary arteries and cerebral vessels, allowing 2D slices to provide sufficient and effective information for segmentation. However, the performance of Baseline2D in vascular connectivity is even poorer (shown in the final row of \figref{pic5_3d_result}\textcolor{subsectioncolor}{(b)} and \textcolor{subsectioncolor}{(e)}) due to the difficulty of focusing on the 3D structural features from slice annotation. \figref{pic6_mip_result} shows the MIP images of testing volumes and their corresponding segmentation results from three datasets. The efficacy of our method can also be seen from the distribution of false negatives and false positives.

\begin{table*}[b]
	\centering
	\scriptsize
	\caption{The effectiveness of confidence learning and uncertainty estimation.  The $p$ of $DSC(p)$ is consistent with \tabref{tab3}.}
	\label{tab6}\renewcommand\arraystretch{1.15}
	\begin{threeparttable}
		\begin{tabular}{ c c c c c c c c c }
			\Xhline{2\arrayrulewidth}
			\multirow{2}*[-2.8pt]{\fontsize{6pt}{7pt} \textbf{Dataset}}
			\rule{0pt}{2pt} &
			\multirow{2}*[-2.8pt]{\fontsize{6pt}{7pt} \textbf{Method}}
			\rule{0pt}{2pt} & \multicolumn{2}{c}{\multirow{1}{*}[-0.6pt]{\fontsize{6pt}{7pt} \textbf{Foreground}}}
			& \multicolumn{2}{c}{\multirow{1}{*}[-0.6pt]{\fontsize{6pt}{7pt} \textbf{Background}}} & \multicolumn{3}{c}{\multirow{1}{*}[-0.6pt]{\fontsize{6pt}{7pt} \textbf{Segmentation Result}}} \\ [-1.5pt]
			\cmidrule(lr){3-4}\cmidrule(lr){5-6}\cmidrule(lr){7-9}
			& & \multirow{1}{*}[0.6pt]{\fontsize{6pt}{7pt}\selectfont \textbf{Num(\%)}}   & \multirow{1}{*}[0.6pt]{\fontsize{6pt}{7pt}\selectfont\textbf{Acc(\%)}} & \multirow{1}{*}[0.8pt]{\fontsize{6pt}{7pt}\selectfont \textbf{Num(\%)}}
			&\multirow{1}{*}[0.8pt]{\fontsize{6pt}{7pt}\selectfont \textbf{Acc(\%)}}
			& \multirow{1}{*}[0.6pt]{\fontsize{6pt}{7pt}\selectfont \textbf{DSC(\%)}} & \multirow{1}{*}[0.8pt]{\fontsize{6pt}{7pt}\selectfont \textbf{ClDice(\%)}} & \multirow{1}{*}[0.6pt]{\fontsize{6pt}{7pt}\selectfont\textbf{AHD(mm)}}   \\
			\hline
			\multirow{4}*[-1.0pt]{\fontsize{6pt}{7pt} \textbf{TubeTK}}
			\rule{0pt}{2pt} & No 
			& 38.31 & 82.47 & 92.94 & 99.86 
			& 60.08 \hspace{0.14cm}($\ast\ast$)\hspace{0.13cm} & 75.25 & 0.971 \\
			& CL
			& 39.19 & 82.59 & 92.92 & 99.87 
			& 60.18 ($\ast\ast\ast$) & 74.99 & 0.866 \\
			& UE  
			& 40.42 & 82.35 & 96.17 & 99.85 
			& 60.64 ($\ast\ast\ast$) & 75.30 & 0.858 \\
			& CL+UE
			& 41.30 & 82.49 & 96.15 & 99.86 
			& 61.10 \hspace{0.23cm}(-)\hspace{0.23cm} & 75.93 & 0.843 \\
			\hline
			\multirow{4}*[-1.0pt]{\fontsize{6pt}{7pt} \textbf{Coronary CTA}}
			\rule{0pt}{2pt} & No
			& 37.14 & 96.05 & 80.16 & 99.97 
			& 74.17 \hspace{0.14cm}($\ast\ast$)\hspace{0.13cm} & 71.10 & 0.699 \\
			& CL
			& 37.52 & 96.39 & 80.14 & 99.98 
			& 75.30 \hspace{0.01cm}(0.115)\hspace{0.01cm} & 72.00 & 0.666\\
			& UE
			& 50.93 & 95.77 & 80.50 & 99.97 
			& 74.56 ($\ast\ast\ast$) & 71.62 & 0.663 \\
			& CL+UE 
			& 51.04 & 96.03 & 80.48 & 99.98
			& 75.43 \hspace{0.23cm}(-)\hspace{0.23cm} & 73.17 & 0.660 \\
			\Xhline{2\arrayrulewidth}
		\end{tabular}
	\end{threeparttable}
\end{table*}

\subsection{Ablation Study}
\label{subsec:ablation}
\subsubsection{The Effectiveness of Each Component}
To verify the efficacy of each component in the proposed framework, we conduct experiments deploying solely 2D features or 3D features, recorded as $L_{\mathrm{2D}}$ and $L_{\mathrm{3D}}$, respectively. Additionally, we present the outcomes without the pseudo-label refinement module ($L_{\mathrm{all}}$). The results (as shown in \tabref{tab5}) demonstrate that relying solely on either 2D or 3D features leads to a degradation in performance, especially when only 2D features are utilized. Our proposed method, which effectively integrates two types of features, achieves better performance. Furthermore, comparison with $L_{\mathrm{all}}$, the framework with enhancing the accuracy of pseudo-labels by confidence learning and uncertainty estimation ($L_{\mathrm{all}}+$RF) achieves superior outcomes across all five datasets.

\subsubsection{Confident Learning and Uncertainty Estimation}
We present an analysis of the performance achieved by incorporating confident learning for the refinement of existing noisy labels ($CL$), as well as incorporating uncertainty estimation to refine unlabeled voxels ($UE$), on two datasets. We conduct an assessment of the quality of generated pseudo labels and the segmentation performance using two optimization strategies in comparison to absence of pseudo-label refinement. A higher quantity ($Num$) and accuracy ($Acc$) of the generated foreground and background labels relative to the real labels is indicative of higher pseudo-label quality. The results, presented in \tabref{tab6}, indicate that both the $CL$ and $UE$ methods effectively enhance label quality. The $UE$ method primarily concentrates on refining unlabeled voxels, leading to a significant impact on the number of voxels in pseudo-label ($Num$). Furthermore, comparison of the results from the two datasets reveals that the addition of background voxels mainly occurs in the TubeTK dataset, while foreground voxels are primarily added in the Coronary CTA dataset. This disparity may stem from the differences in annotation quality between the two datasets, with the TubeTK dataset exhibiting inherent noise (some venous) in its annotations, resulting in lower prediction uncertainty for the background. Moreover, we identify a positive correlation between the quality of generated pseudo labels and the final segmentation results, in line with our expectations. And comparative analysis against pseudo-labels generated solely through traditional methods ($No$) reveals an improvement in segmentation performance of network upon both these two strategies. And the combined utilization of confident learning and uncertainty estimation yields best outcomes in the network's performance.

\subsubsection{The Parameter of Pseudo-Label Refinement}
\begin{figure}[!h]
	\captionsetup[subfigure]{font={sf,scriptsize}}
	\subfloat[\normalfont{Foreground}]{\includegraphics[width=.46\linewidth]{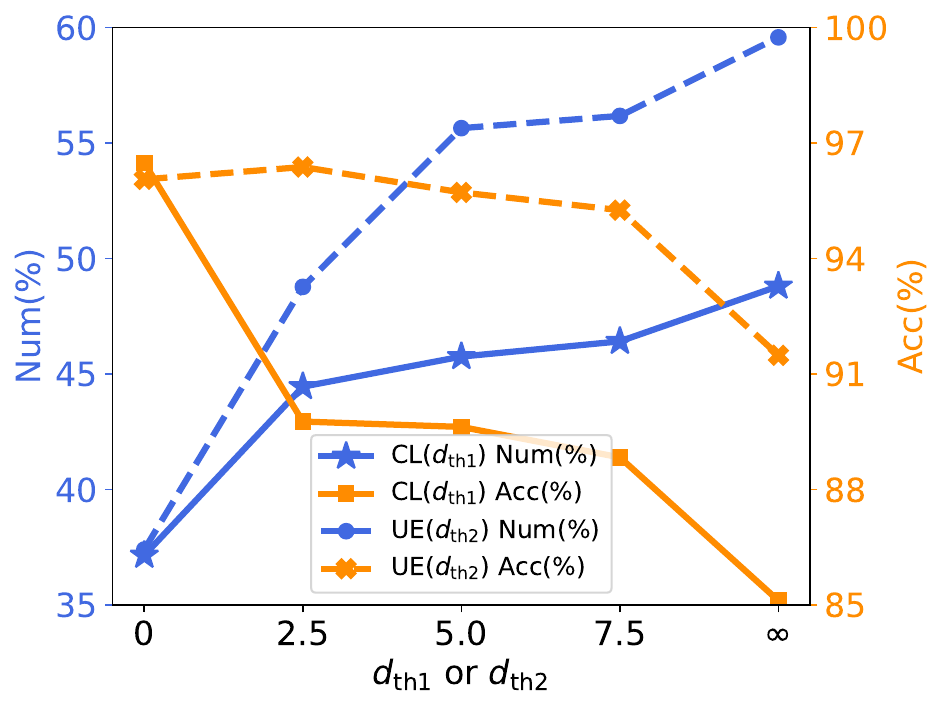}
		\label{pic9_a}%
	}
	\hfil
	\subfloat[\normalfont{Background}]{\includegraphics[width=.48\linewidth]{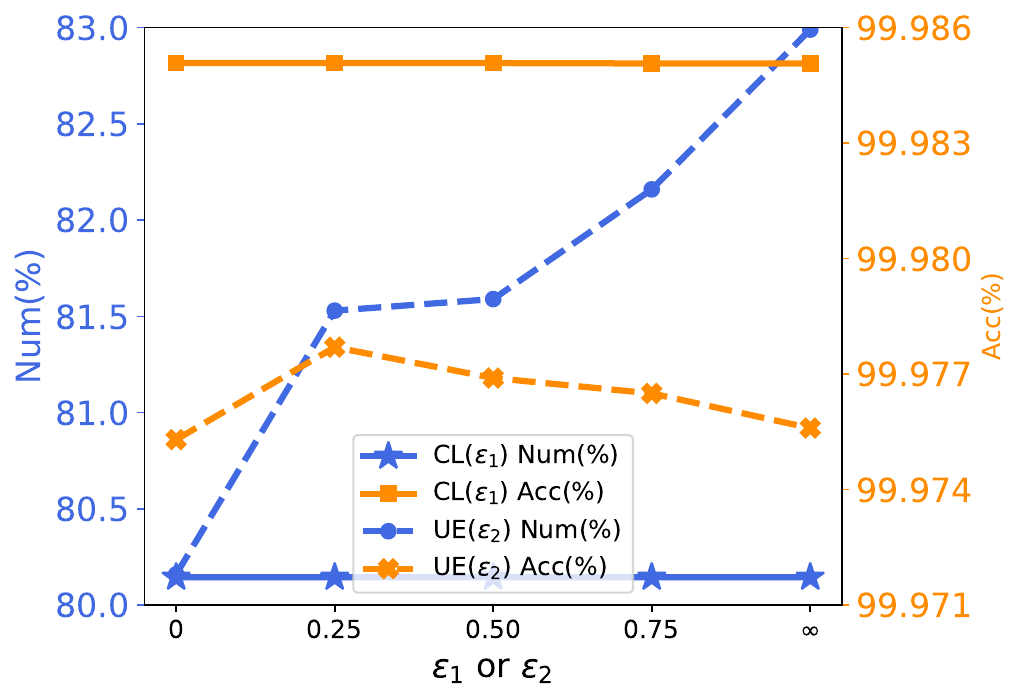}\label{pic9_b}%
	}
	\caption{\protect\subref{pic9_a} shows the impact of $d_{\mathrm{th1}}$ and $d_{\mathrm{th2}}$ on the generated foreground, and \protect\subref{pic9_b} is the impact of $\varepsilon_1$ and $\varepsilon_2$ on the background.}
	\label{pic7_param}
\end{figure}

The control over foreground generation primarily rests with parameters $d_{\mathrm{th1}}$ and $d_{\mathrm{th2}}$, while background generation is primarily regulated by $\varepsilon_1$ and $\varepsilon_2$ (\textcolor{subsectioncolor}{Eq.} \ref{CL_computing} and \textcolor{subsectioncolor}{Eq.} \ref{Un_computing}). To explore the impact of parameter variations on the generation of foreground and background, we conduct experiments on the Coronary CTA dataset, as depicted in \figref{pic7_param}. The \textit{Num} metric indicates the proportion of the number of generated voxels to the actual number, while \textit{Acc} represents the accuracy rate. Within the confidence learning module, increasing the value of $d_{\mathrm{th1}}$ results in the inclusion of more foreground voxels from $S_0^{\mathrm{(re)}}$. However, this increment is accompanied by a decline in $Acc$. Similarly, within the uncertainty estimation module, increasing $d_{\mathrm{th2}}$ leads to an increased allocation of foreground voxels to unlabeled voxels and a decrease in accuracy. Regarding background voxels, the variation of $\varepsilon_1$ has minimal observable impact on $Acc$ and $Num$ of generated background due to $|S_1^{\mathrm{(re)}}| \ll |S_0|$. And similar to $d_{\mathrm{th2}}$, as $\varepsilon_2$ increases, the number of allocated background voxels increases while the accuracy declines. The observed variations align with our initial expectations. The numerical selection of $d_{\mathrm{th1}}$ and $d_{\mathrm{th2}}$ aims to strike a balance between the quantity and accuracy of the generated foreground set. And these adjustments additionally impact the subsequent fine-tuning process of the segmentation network. Similar considerations apply to $\varepsilon_1$ and $\varepsilon_2$. During the experimental analysis, we maintain consistent parameter settings across the four datasets, with the exception of TubeTK dataset. This consistent performance demonstrates the robust generalization capabilities of our proposed method.

\subsection{Generalization of Method}

\begin{table}[!h]
	\centering
	\scriptsize
	\caption{Cross-validation on TubeTK and Cerebral CTA datasets. The $p$ of $DSC(p)$ is consistent with \tabref{tab3}.}
	\label{tab-cross}\renewcommand\arraystretch{1.15}
	\begin{threeparttable}
		\begin{tabular}{ c c c c c }
			\Xhline{2\arrayrulewidth}
			\multirow{2}*[-2.8pt]{\fontsize{6pt}{7pt} \textbf{Method}}
			\rule{0pt}{2pt} & \multicolumn{2}{c}{\multirow{1}{*}[-0.6pt]{\fontsize{6pt}{7pt} \textbf{TubeTK}}}
			& \multicolumn{2}{c}{\multirow{1}{*}[-0.6pt]{\fontsize{6pt}{7pt} \textbf{Cerebral CTA}}} \\ [-1.5pt]
			\cmidrule(lr){2-3}\cmidrule(lr){4-5}
			& \multirow{1}{*}[0.6pt]{\fontsize{6pt}{7pt}\selectfont \textbf{DSC(\%)}}   & \multirow{1}{*}[0.6pt]{\fontsize{6pt}{7pt}\selectfont\textbf{AHD(mm)}} 
			& \multirow{1}{*}[0.6pt]{\fontsize{6pt}{7pt}\selectfont \textbf{DSC(\%)}} & \multirow{1}{*}[0.6pt]{\fontsize{6pt}{7pt}\selectfont\textbf{AHD(mm)}}   \\
			\hline
			Full-sup 
			& 63.90 ($\ast\ast\ast$) & 0.998
			& 84.96 ($\ast\ast\ast$) & 0.387\\
			\hdashline
			Baseline3D
			& 56.85 ($\ast\ast\ast$) & 1.387 
			& 78.72 ($\ast\ast\ast$) & 0.543 \\
			Baseline2D  
			& 59.40 ($\ast\ast\ast$) & 1.035 
			& 81.86 ($\ast\ast\ast$) & 0.488 \\
			MTCL
			& 60.05 \hspace{0.25cm}($\ast$)\hspace{0.19cm} & 1.008
			& 74.75 ($\ast\ast\ast$) & 1.347 \\
			SLD
			& 58.18 ($\ast\ast\ast$) & 1.307 
			& 81.49 ($\ast\ast\ast$) & 0.645 \\
			SPDS
			& 52.53 ($\ast\ast\ast$) & 1.602 
			& 78.15 ($\ast\ast\ast$) & 0.459 \\
			Ours 
			& \textbf{60.38} \hspace{0.24cm}(-)\hspace{0.24cm} & \textbf{0.936} 
			& \textbf{84.96} \hspace{0.22cm}(-)\hspace{0.23cm} & \textbf{0.297} \\
			\Xhline{2\arrayrulewidth}
		\end{tabular}
	\end{threeparttable}
\end{table}

\subsubsection{Cross-Validation Experiments} To further demonstrate the generalization of our method, we conduct cross-validation experiments on two relatively small datasets, as presented in \tabref{tab-cross}. By comparing the results, we can draw similar conclusions to those observed from \tabref{tab3} and \tabref{tab4}: our method effectively utilizes the carefully designed weak labels and achieves superior performance compared to other methods within similar annotation timeframes. Moreover, in the cross-validation experiments, the entire dataset is used for testing, resulting in more robust and stable outcomes. Consequently, compared to the results in \tabref{tab3}, our method exhibits higher statistical significance when evaluated against MTCL on the TubeTK dataset. This observation further supports that our approach can deliver superior results compared to MTCL on datasets with noisy labels.

\subsubsection{Robustness} 
\begin{figure}[h]
	\captionsetup[subfigure]{font={sf,scriptsize}}
	\centering
	\subfloat[\normalfont{Cerebral CTA}]{\includegraphics[width=.46\linewidth]{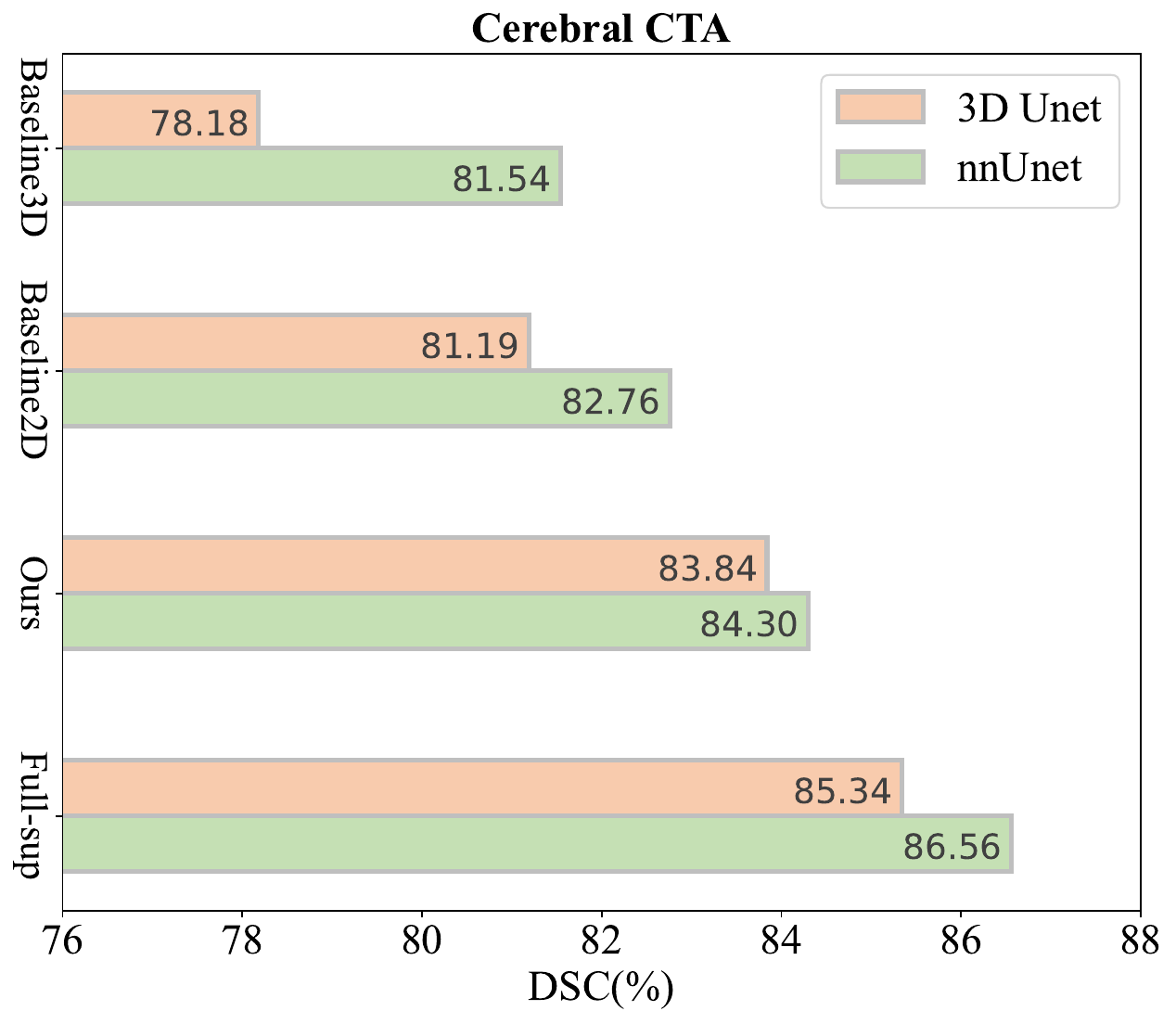}
		\label{pic_backbobe_a}%
	}
	\hfil
	\subfloat[\normalfont{Coronary CTA}]{\includegraphics[width=.46\linewidth]{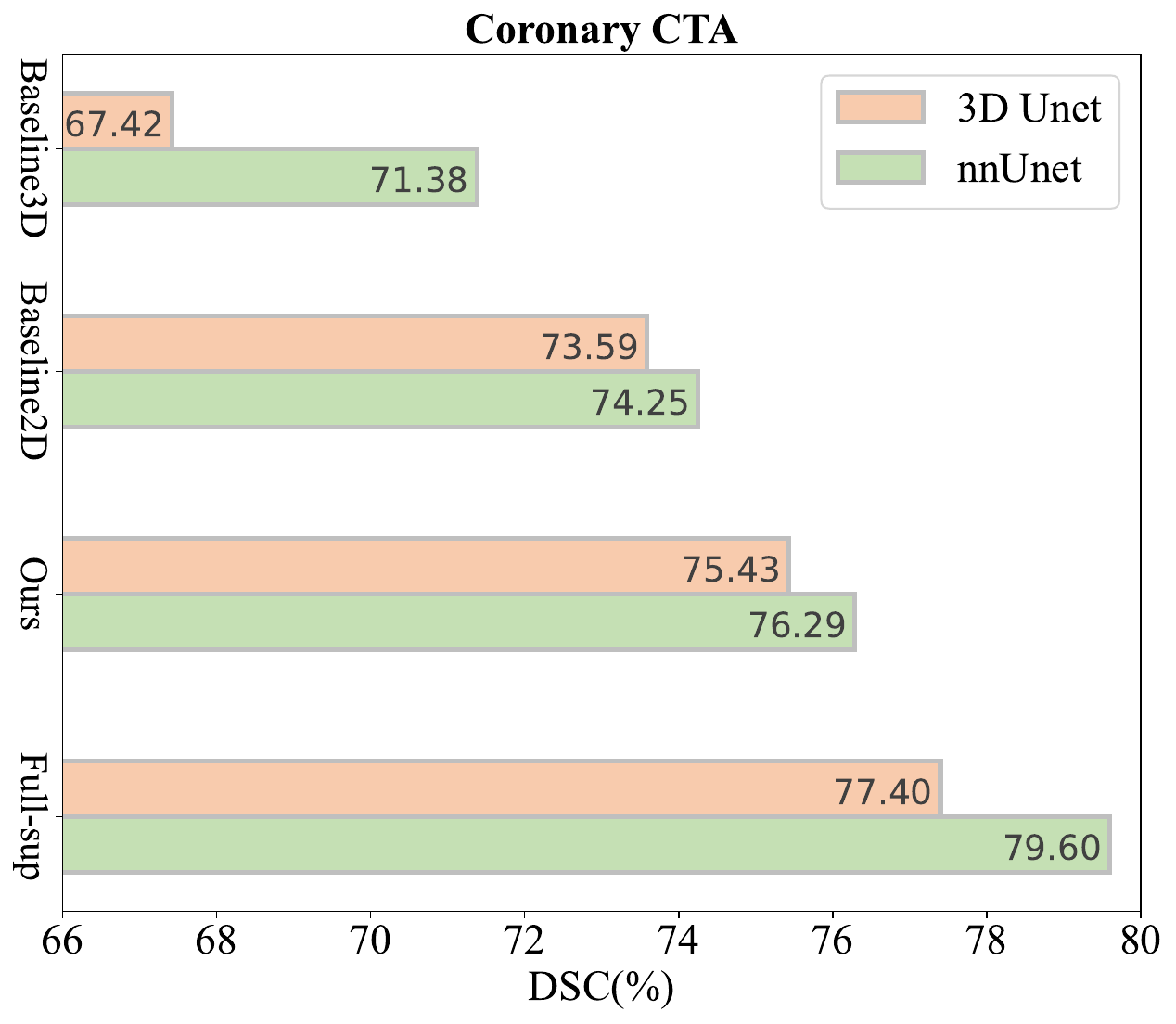}\label{pic_backbobe_b}%
	}
	\caption{DSC metric results under different backbones on two datasets.}
	\label{pic8_backbone}
\end{figure}

In this subsection, we primarily focus on the robustness of our proposed weakly-supervised segmentation framework. Our scheme can leverage most fully supervised methods as backbones, allowing us to analyze their impact on the segmentation performance. Due to the superior performance, nnUnet \cite{isensee2021nnu} is widely used in various medical image segmentation tasks. \figref{pic8_backbone} illustrates the results achieved by various methods using 3D Unet and nnUnet as backbones, respectively. Notably, we utilize initially generated pseudo labels ($S_0,$ $S_1$) to obtain the `data fingerprint' and `pipeline fingerprint' (design parameters of nnUnet) of the backbone for proposed framework when employing nnUnet. Based on the findings presented in \figref{pic8_backbone}, it is evident that the methods exhibit similar characteristics under both backbones. And with the same backbone (3D Unet or nnUnet), our approach effectively harnesses the labels of MIP images, yielding superior results compared to Baseline2D and Baseline3D, while approaching the performance of fully supervised methods. Furthermore, a horizontal comparison between the two backbones reveals that nnUnet outperforms 3D Unet, aligning with expectations due to nnUnet's ability to fully exploit the dataset characteristics.

\subsection{Trade-off Between Annotation Time and Performance}
\label{subsec:trade-off}

\subsubsection{Further Optimization of Annotation Method}

\label{subsubsec:Further}
\begin{figure}[h]
	\captionsetup[subfigure]{font={scriptsize}}
	\centering
	\subfloat[\normalfont{MIP image}]{\includegraphics[width=.25\linewidth]{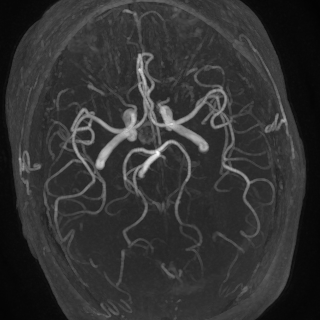}\label{pic_mip_a}%
	}
	\hfil
	\subfloat[\normalfont{Manual annotation}]{\includegraphics[width=.25\linewidth]{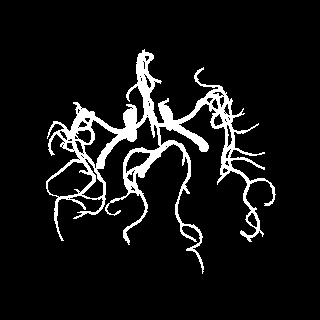}\label{pic_mip_b}%
	}
	\hfil
	\subfloat[\normalfont{RN}]{\includegraphics[width=.25\linewidth]{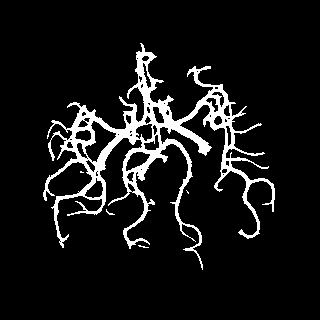}\label{pic_mip_c}%
	}\\
	
	\subfloat[\normalfont{RN+TV}]{\includegraphics[width=.25\linewidth]{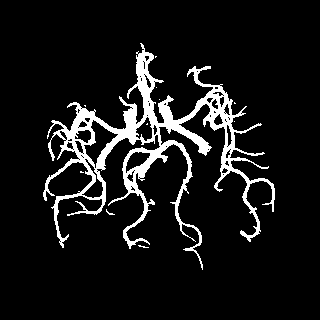}\label{pic_mip_d}%
	}
	\hfil
	\subfloat[\normalfont{Num(10\%)}]{\includegraphics[width=.25\linewidth]{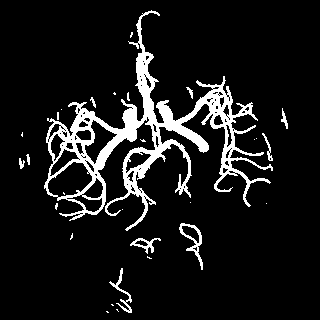}\label{pic_mip_e}%
	}
	\hfil
	\subfloat[\normalfont{Num(30\%)}]{\includegraphics[width=.25\linewidth]{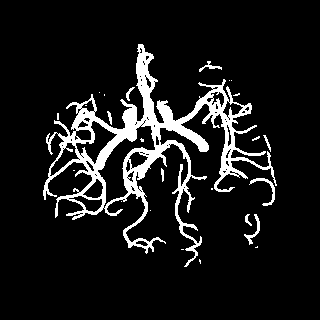}\label{pic_mip_f}%
	}
	\caption{Different annotations of one MIP image, where \protect\subref{pic_mip_b} is the label with completely manual annotation. And \protect\subref{pic_mip_e} and \protect\subref{pic_mip_f} are the predicted results of 2D Unet trained with the annotated MIP images (Num(10\%), Num(30\%)).}
	\label{pic_mip_annotation}
\end{figure}

In our method, the annotation of 2D projected blood vessels is an essential aspect. Despite being simpler than annotating in 3D space, it necessitates manual intervention by radiologists, with an average annotation time per image of 6-7 minutes (shown in \tabref{tab2}). To segment 2D vascular images, conventional methods or deep learning methods can be applied for automatic or semi-automatic segmentation. Although these approaches may impact annotation accuracy, it offers significant reduction of annotation workload and potential for unsupervised and semi-supervised segmentation. In this section, we investigate the impact of reducing annotation time on results by employing both a conventional method and a learning-based approach on Cerebral MRA dataset (consistent with \tabref{tab2}). We combine Frangi filtering \cite{frangi1998multiscale} and homomorphic filtering techniques for the segmentation of MIP images, followed by the manual removing of obvious noise (RN) and labeling of thick vessels (TV) to acquire MIP annotations of varying qualities. These results are then integrated into our framework as 3D weak labels to derive the final 3D segmentation result. Furthermore, varying numbers (10\%, 30\%) of MIP images from training set are annotated, allowing us to train a 2D Unet \cite{ronneberger2015u} for obtaining segmentation results of the remaining MIP images. Subsequently, the manually annotated labels and the segmentation results serve as weakly supervised labels for training our proposed network. \figref{pic_mip_annotation} presents the annotated images obtained through different methods.

\begin{table*}[b]
	\centering
	\scriptsize
	\caption{The impact of different annotation methods on the segmentation results. The $p$ of $DSC(p)$ is consistent with \tabref{tab3}. When assessing the quality of MIP annotation in 2D Unet method, we concurrently evaluate the manually annotated training data (Dice=100\%, Acc=100\%) and the generated segmentation results, as they collectively serve as weak labels for our method, facilitating easier comparison with Frangi+Homomorphic.}
	\label{tab:Further-Optimization}\renewcommand\arraystretch{1.15}
	\begin{threeparttable}
		\begin{tabular}{ c c c c c c c c c }
			\Xhline{2\arrayrulewidth}
			\multirow{2}*[-2.8pt]{\fontsize{6pt}{7pt} \textbf{Mehod}}
			\rule{0pt}{2pt} &
			\multirow{2}*[-2.8pt]{\fontsize{6pt}{7pt} \textbf{Scheme}}
			\rule{0pt}{2pt} & \multicolumn{4}{c}{\multirow{1}{*}[-0.6pt]{\fontsize{6pt}{7pt} \textbf{MIP Annotation}}}
			& \multicolumn{3}{c}{\multirow{1}{*}[-0.6pt]{\fontsize{6pt}{7pt} \textbf{Segmentation Result}}}  \\ [-1.5pt]
			\cmidrule(lr){3-6}\cmidrule(lr){7-9}
			& & \multirow{1}{*}[0.6pt]{\fontsize{6pt}{7pt}\selectfont \textbf{Ave(min)}}   & \multirow{1}{*}[0.6pt]{\fontsize{6pt}{7pt}\selectfont\textbf{Dice(\%)}} & \multirow{1}{*}[0.8pt]{\fontsize{6pt}{7pt}\selectfont \textbf{Acc(\%)}} &\multirow{1}{*}[0.8pt]{\fontsize{6pt}{7pt}\selectfont \textbf{Example}}
			&\multirow{1}{*}[0.8pt]{\fontsize{6pt}{7pt}\selectfont \textbf{Dice(\%)}}
			& \multirow{1}{*}[0.8pt]{\fontsize{6pt}{7pt}\selectfont \textbf{ClDice(\%)}} & \multirow{1}{*}[0.6pt]{\fontsize{6pt}{7pt}\selectfont\textbf{AHD(mm)}}   \\
			\hline
			\multirow{2}*[-1.0pt]{\fontsize{6pt}{7pt} \shortstack{\textbf{Frangi+Homomorphic}}}
			& RN
			& 0.86 & 75.83 & 95.75 & \figref{pic_mip_annotation}\subref{pic_mip_c}
			& 73.84 ($\ast\ast\ast$) & 64.48 & 1.282 \\
			& RN+TV  
			& 2.39 & 86.13 & 97.41 & \figref{pic_mip_annotation}\subref{pic_mip_d}
			& 79.79 ($\ast\ast\ast$) & 77.56 & 0.830 \\
			\hdashline
			\multirow{2}*[-1.0pt]{\fontsize{6pt}{7pt} \textbf{2D Unet}}
			\rule{0pt}{2pt} & Num(10\%)
			& 0.68 & 81.35 & 96.36 & \figref{pic_mip_annotation}\subref{pic_mip_e} and \subref{pic_mip_b}
			& 78.85 ($\ast\ast\ast$) & 79.51 & 1.191 \\
			& Num(30\%)
			& 2.26 & 87.73 & 97.66 & \figref{pic_mip_annotation}\subref{pic_mip_f} and \subref{pic_mip_b}
			& 82.42 ($\ast\ast\ast$) & 84.36 & 0.509\\
			\hdashline
			\textbf{Mannually} & - 
			& 6.78 & 100 & 100 & \figref{pic_mip_annotation}\subref{pic_mip_b}
			& 84.35 \hspace{0.24cm}(-)\hspace{0.23cm} & 87.40 & 0.336 \\
			\Xhline{2\arrayrulewidth}
		\end{tabular}
	\end{threeparttable}
\end{table*}

\tabref{tab:Further-Optimization} presents the average labeling time required for weakly supervised labels obtained through various methods, in addition to the MIP labeling quality (Dice coefficient and accuracy) and their implications on the final segmentation results. Initially, as expected, it is observed that modifying all or part of the manual annotations by incorporating algorithm-generated pseudo labels leads to a reduction in annotation time, resulting in a decrease in the quality of MIP annotation and a negative impact on segmentation results. Furthermore, the comparison of the results obtained through two different methods reveals that, when utilizing 2D Unet, the quality of MIP annotations and the performance of the final segmentation results outperform those of traditional method under similar annotation time. This can be attributed to the inclusion of manually obtained correct labels in the 2D Unet method, enabling the network to assimilate more valuable information during the learning process. Additionally, the comparison between the results of $Num(30\%)$ and all manual annotations ($Manually$) indicates that employing 2D Unet can significantly reduce annotation time with only a slight decrease in segmentation performance. This highlights the potential of our framework to integrate with other 2D segmentation methods for further reduction of the annotation workload without significant performance compromise, a topic to be further discussed in Sec. \ref{subsec:limitation}.

\subsubsection{The Number of Weakly Annotations}
\begin{figure}[!h]
	\captionsetup[subfigure]{font={scriptsize}}
	\centering
	\subfloat[\normalfont{Labeled samples}]{\includegraphics[width=.48\linewidth]{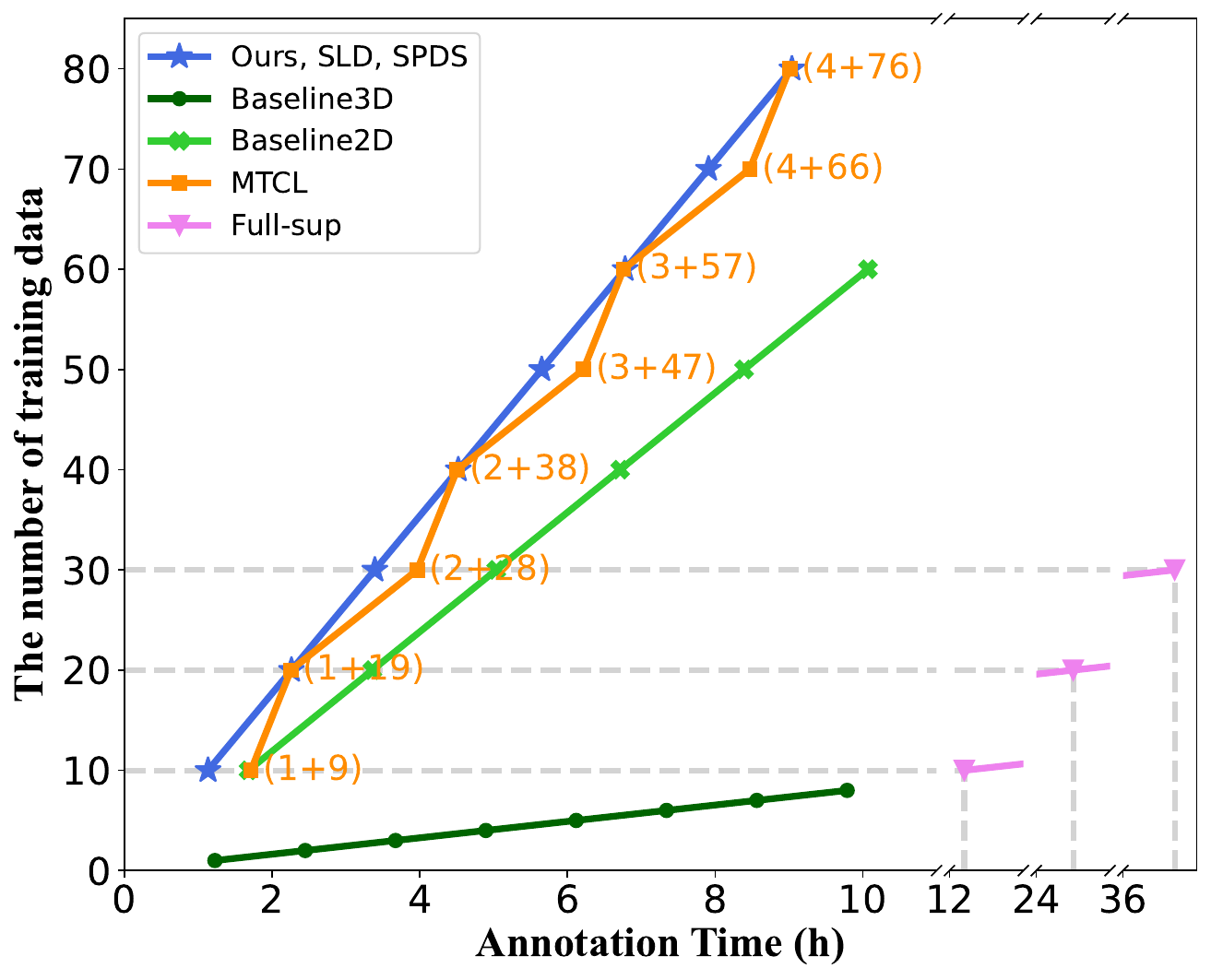}\label{pic8_a}%
	}
	\hfil
	\subfloat[\normalfont{Dice}]{\includegraphics[width=.48\linewidth]{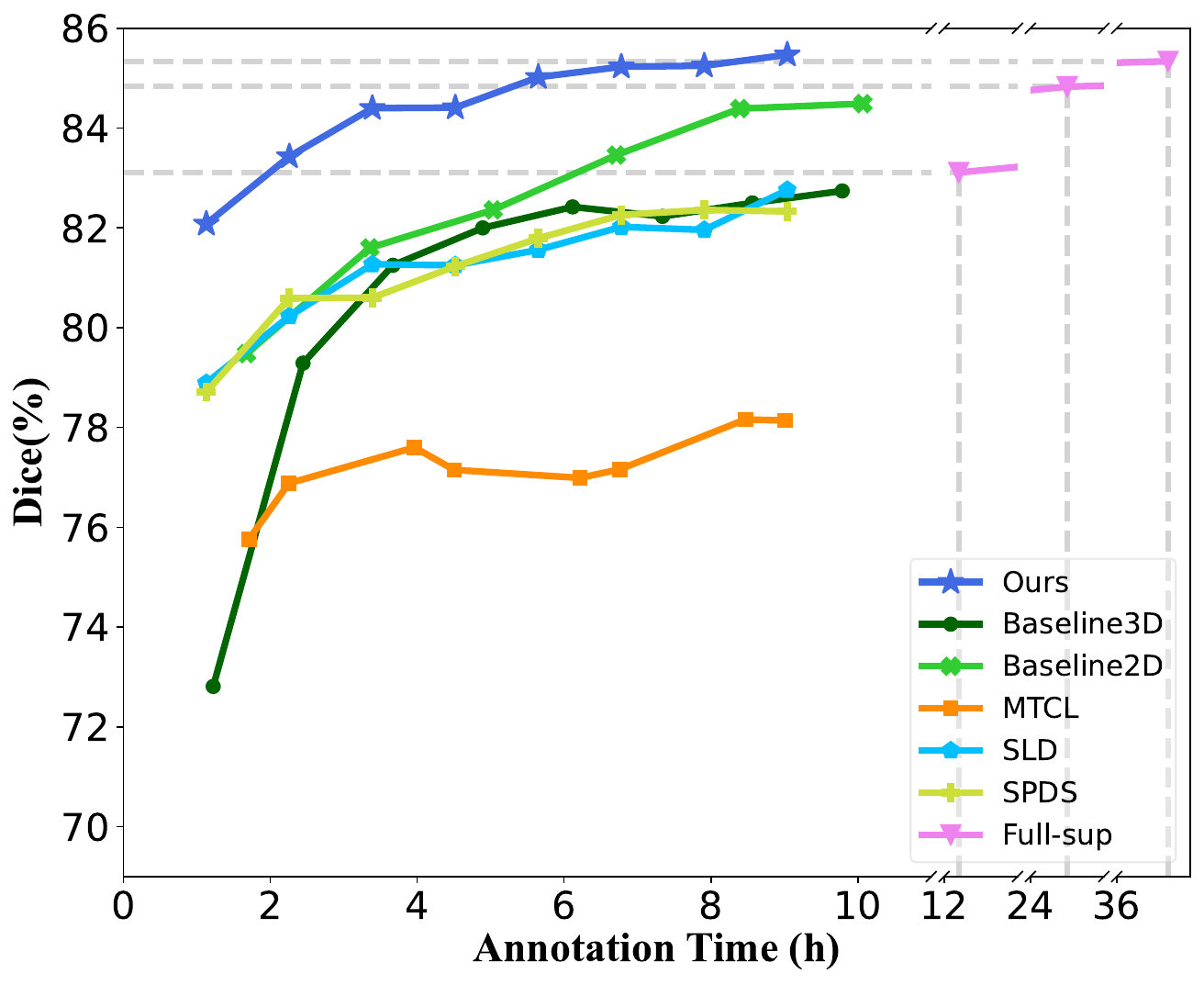}\label{pic8_b}%
	}\\
	
	\subfloat[\normalfont{ClDice}]{\includegraphics[width=.48\linewidth]{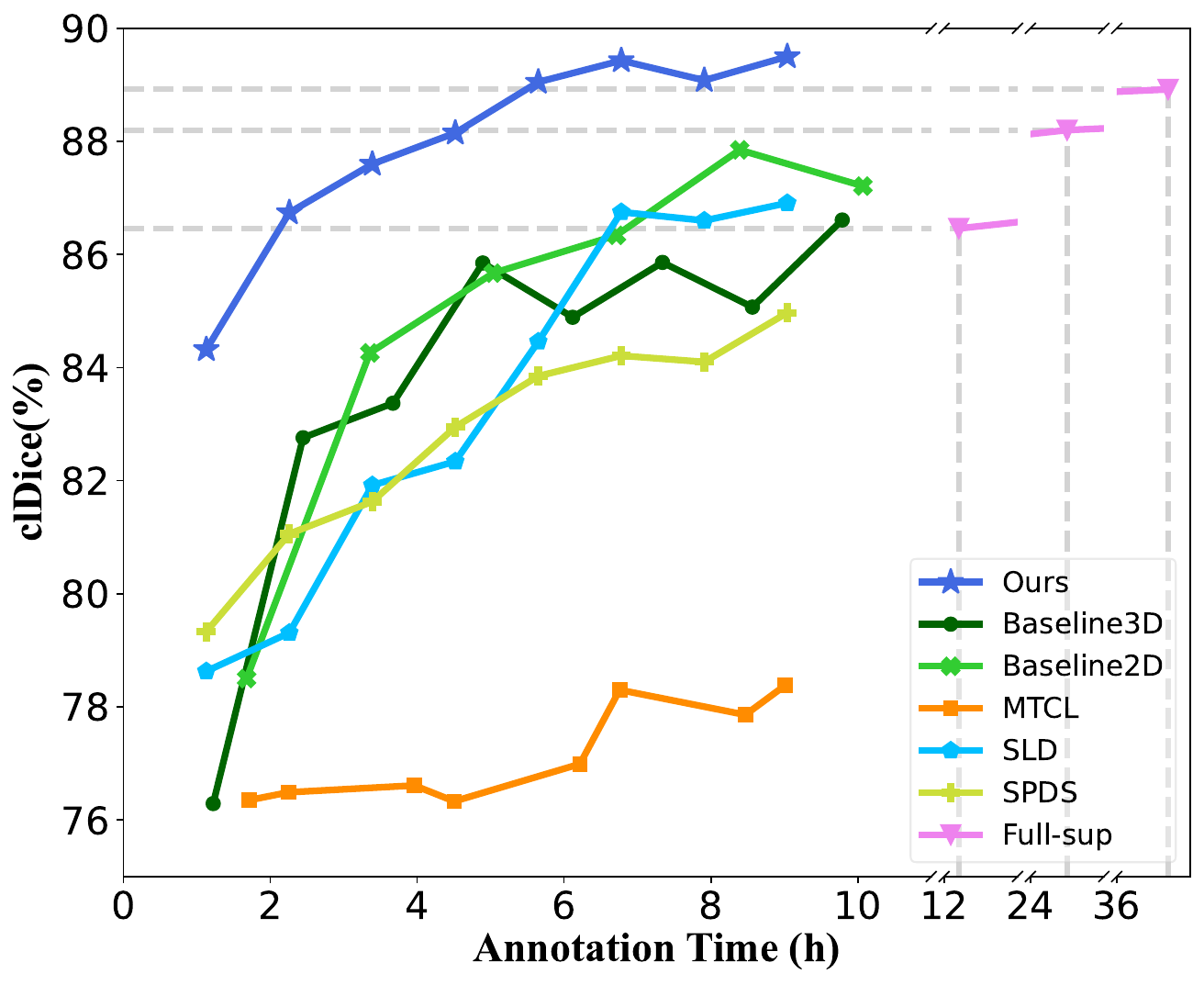}\label{pic8_c}%
	}
	\hfil
	\subfloat[\normalfont{AHD}]{\includegraphics[width=.48\linewidth]{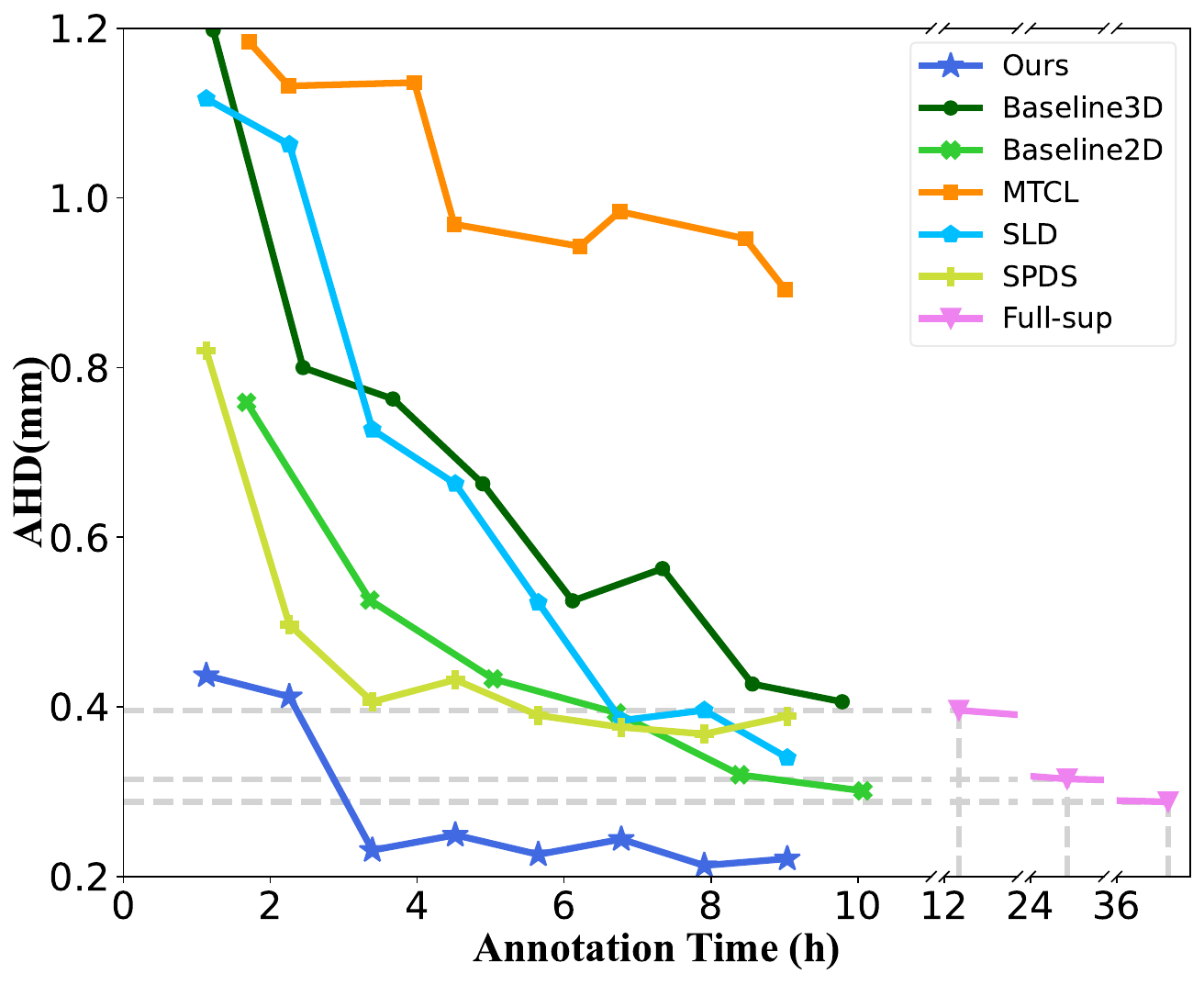}\label{pic8_d}%
	}
	\caption{\protect\subref{pic8_a} shows the number of training images and annotation time of each method, where the numbers of full annotation and noisy label are shown for MTCL. Notably, the lines of SLD and SPDS in \protect\subref{pic8_a} align with our method, as identical training labels are utilized to ensure fair comparison (mentioned in Sec. \ref{subsec:methodtime}). \protect\subref{pic8_b}-\protect\subref{pic8_d} show the segmentation performance vs. annotation time. Notably, the annotation time of fully supervised method (Full-sup) trained with the fewest number of samples (12.2h, 10 training samples) is longer than that of the proposed method (Ours) trained with the greatest number of samples (9.0h, 80 training samples), which is why there is no overlap between the two methods on the axis of Annotation Time.}
	\label{pic10_trade_off}
\end{figure}

The performance of weakly-supervised method theoretically does not exceed that of fully supervised learning under the same amount of training data. However, it is meaningful to study whether weakly-supervised method can outperform fully supervised method in less annotation time by adding weakly labeled training samples. This area of research has received limited attention in prior studies on weakly supervised learning. In our study, we randomly select 12 samples as testing set and 4 samples as the validation set from the 96 volumes of Cerebral MRA dataset. In \figref{pic10_trade_off}, we present the results of monitoring the performance trends of each method under varying labeling times (different numbers of training samples) on Cerebral MRA dataset, with the same testing volumes. Additionally, we evaluate the performance of full supervision with 10, 20, and 30 training samples which require labeling times of 12.2h, 24.5h, and 36.7h respectively. 

Based on the results presented in \figref{pic10_trade_off}, it is evident that the performance of each method shows a gradual increase with the expansion of training data, consistent with anticipated outcomes. And our method outperforms other methods across all indicators in the case where the annotation time of training data is similar. Furthermore, when the number of weakly-labeled samples is sufficient, our proposed method can outperform full supervision while still requiring far less labeling time. For example, our approach achieves better results than fully supervised segmentation (trained on 30 images) with only about 7.9h of data annotation time, which is significantly less than the 36.7h required for the latter. Similarly, in much less annotation time (about 2.3h, 5.7h), our method outperforms the performance under full supervision with the labeling time of 12.2h and 24.5h.

\subsection{Limitation and Future Works}
\label{subsec:limitation}
One limitation of our study pertains to the impact of blood vessel occlusion on the accurate labeling of the MIP image. Our proposed methodology necessitates the blood vessel annotation of the MIP image. But the presence of diverse types of blood vessels introduces varying degrees of occlusion challenges due to dissimilar surrounding tissues. Consequently, some preprocessing procedures may be required for certain datasets (e.g., the Aorta CTA dataset and the Coronary CTA dataset) in the training phase. Furthermore, occlusion is present in the images of the patients who undergo surgery and implant metal materials, which is not considered in our work. Hence, it is intriguing to investigate how to attain superior performance in the presence of occlusion, even when it is severe.

Additionally, our approach mandates the annotation of 2D MIP blood vessel images, which is still time-consuming. In Sec. \ref{subsubsec:Further}, we attempt to replace manual full annotation of MIP images with automatic methods (with minor manual annotation); however, these methods yield unsatisfactory results, introducing noise that adversely affects the final segmentation outcome. Exploring the utilization of existing 2D blood vessel weakly-supervised and semi-supervised methods to minimize the annotation workload while upholding accurate blood vessel segmentation represents a promising avenue for future research.

\section{Conclusion}
\label{sec:conclusion}

In this study, we present a framework for weakly supervised segmentation of vessels in 3D volumes with dimensionality reduction annotation, leveraging the sparse structural characteristics of vascular structure. To this end, MIP image is employed as a means of annotation and supervision. Initially, we obtain pseudo-label of 3D vessels through MIP annotation. Subsequently, we design a 2D-3D feature fusion network to make best use of the weak label, taking into account the acquisition method of the 2D labels. During the pseudo-label generation, it is inevitable that some noise is introduced and certain voxels may be overlooked. To mitigate these issues and enhance network performance, we integrate confidence learning and uncertainty estimation methods to refine the pseudo-labels. We conduct comprehensive experiments across five vascular datasets. And the results demonstrate that our proposed method achieves high-quality vascular segmentation, approaching the performance of fully-supervised segmentation under the same number of training samples. Furthermore, we design experiments to validate that our proposed weakly supervised segmentation framework achieves superior performance to fully supervised segmentation with much less annotation time by increasing training samples, showing the immense potential of our method in the field of vessel segmentation.

\makeatletter
\ifvoid\@mpfootins\else\unvbox\@mpfootins\fi
\makeatother

\end{document}